\pdfoutput=1

\documentclass[11pt]{article}
\usepackage[table]{xcolor}

\usepackage[]{EMNLP2022}

\usepackage{times}
\usepackage{latexsym}

\usepackage[T1]{fontenc}

\usepackage[utf8]{inputenc}

\usepackage{microtype}

\usepackage{inconsolata}

\usepackage{times}
\usepackage{latexsym}

\usepackage{pbox}
\usepackage{makecell}
\usepackage{graphicx}
\usepackage{subcaption}
\usepackage{setspace}
\usepackage{enumitem}
\usepackage{booktabs, cellspace, multirow}
\usepackage{amsmath}
\usepackage{amssymb}
\usepackage{soul}
\usepackage{xspace}
\usepackage{caption}
\usepackage{tabularx}
\usepackage{verbatim}
\usepackage{float}
\usepackage{multirow}
\usepackage{arydshln}
\usepackage{color, colortbl}

\newcommand{\capdataname}{InstructDial}
\newcommand{\dataname}{\textsc{InstructDial}\xspace}
\newcommand{\modelnamebase}{DIAL}
\newcommand{\modelnamebartzero}{\modelnamebase{-BART0}\xspace}
\newcommand{\modelnametzero}{\modelnamebase{-T0}\xspace}

\newcommand{\dataurl}{https://github.com/prakharguptaz/Instructdial}
\definecolor{Gray}{gray}{0.9}
\usepackage{fix-cm}

\title{\capdataname: Improving Zero and Few-shot Generalization in Dialogue through Instruction Tuning}

\author{Prakhar Gupta$^\clubsuit$ \quad Cathy Jiao$^\clubsuit$ \quad Yi-Ting Yeh$^\clubsuit$ \quad Shikib Mehri$^\clubsuit$ \\ \quad \textbf{Maxine Eskenazi}$^\clubsuit$ 
\quad \textbf{Jeffrey P. Bigham}$^{\clubsuit,\heartsuit}$  \\
$^\clubsuit$Language Technologies Institute, Carnegie Mellon University \\
$^\heartsuit$Human-Computer Interaction Institute, Carnegie Mellon University \\
\texttt{\small \{prakharg,cljiao,yitingye,amehri,max,jbigham\}@cs.cmu.edu}}

\begin{document}
\maketitle

\begin{abstract}
Instruction tuning is an emergent paradigm in NLP wherein natural language instructions are leveraged with language models to induce zero-shot performance on unseen tasks.
Dialogue is an especially interesting area in which to explore instruction tuning because dialogue systems perform multiple tasks related to language (e.g., natural language understanding and generation, domain-specific interaction), yet
instruction tuning has not been systematically explored for dialogue-related tasks. 
We introduce \dataname, an instruction tuning framework for dialogue, which consists of a repository of 48 diverse dialogue tasks in a unified text-to-text format created from 59 openly available dialogue datasets.
We explore cross-task generalization ability on models tuned on \dataname across diverse dialogue tasks.
Our analysis reveals that \dataname enables good zero-shot performance on unseen datasets and tasks such as dialogue evaluation and intent detection, and even better performance in a few-shot setting. %
To ensure that models adhere to instructions, we introduce novel meta-tasks. We establish benchmark zero-shot and few-shot performance of models trained using the proposed framework on multiple dialogue tasks\footnote{Code available at \url{\dataurl}}.

\end{abstract}

\section{Introduction}
\label{sec:introduction}

Pretrained large language models (LLMs) ~\cite{devlin-etal-2019-bert, Radford2019LanguageMA, brown2020language} 
are not only few-shot learners, but can also perform numerous language tasks without the need for fine-tuning.
However, LLMs are expensive to train and test.
Instruction tuning has emerged as a tool for directly inducing zero-shot generalization on unseen tasks in language models 
by using natural language instructions ~\cite{mishra2021natural,sanh:iclr22,flan, ouyang2022training}. 
Natural language instructions can contain components such as task definitions, examples, and prompts which allows them to be customized for multitask learning.
Instruction tuning enables developers, practitioners, and even non-expert users to leverage language models for novel tasks by specifying them through natural language, without the need for large training datasets.
Furthermore, instruction tuning can work for models that are significantly smaller than LLMs~\cite{mishra2021natural, sanh:iclr22}, making them more practical and affordable. 

\begin{figure}[t]
\centering
\includegraphics[width=0.42\textwidth]{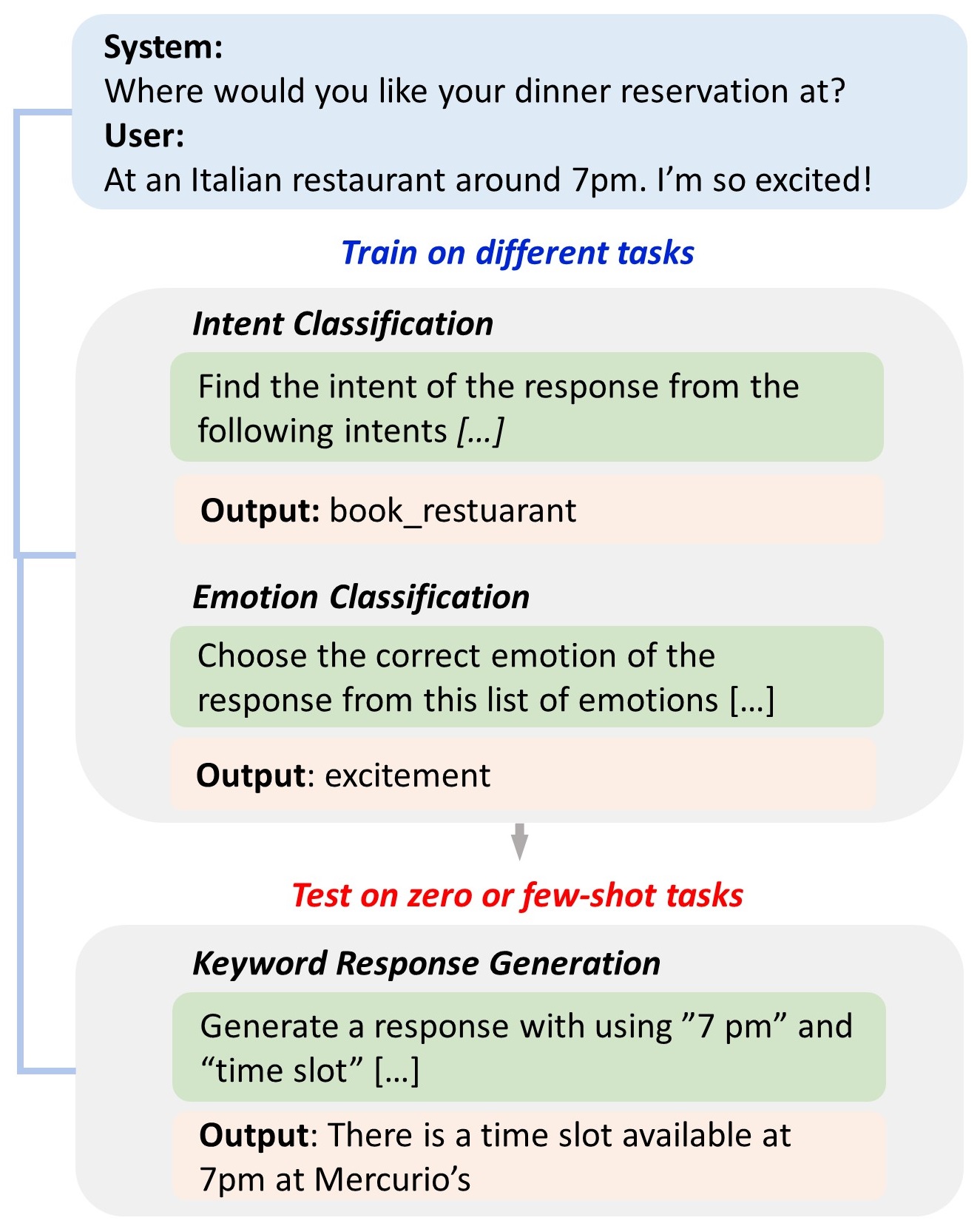}
\caption{We investigate instruction tuning on dialogue tasks. Instruction tuning involves training a model on a mixture of tasks defined through natural language instructions. Instruction tuned models exhibit zero-shot or few-shot generalization to new tasks. 
}
\vspace{-2mm}
\label{fig:intro}
\end{figure}

\begin{figure*}[!htb]
\centering
\includegraphics[width=0.9\textwidth]{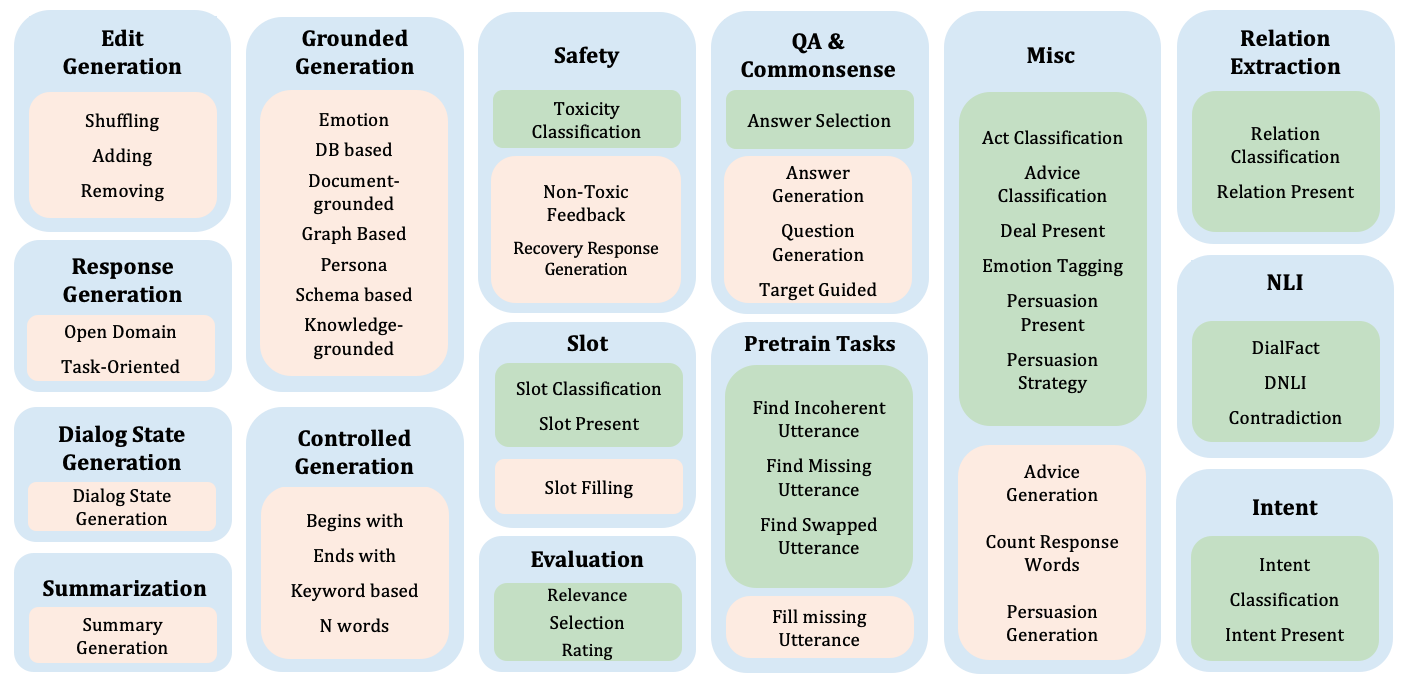}
\caption{\dataname task taxonomy. Green represents classification and orange represents generation tasks.}
\label{fig:tasks}
\vspace{-1em}
\end{figure*}

Most recent work~\cite{mishra2021natural,sanh:iclr22,flan} on instruction tuning has focused on general NLP tasks such as paraphrase detection and reading comprehension, but not specifically on dialogue. 
While some work such as  ~\cite{wang2022benchmarking} include a few dialogue tasks, those tasks are collected through crowdsourcing and do not provide good coverage of dialogue tasks and domains.
No prior work has examined how training a model on a wide range of dialogue tasks with a variety of instructions may affect a system's ability to perform on both core dialogue tasks such as intent detection and response generation, and domain-specific tasks such as emotion classification. In this work, we introduce \dataname, a framework for instruction tuning on dialogue tasks. We provide a large curated collection of 59 dialogue datasets and 48 tasks, benchmark models, and a suite of metrics for testing the zero-shot and few-shot capabilities of the models. \dataname consists of multiple dialogue tasks converted into a text-to-text format (Figure \ref{fig:intro}). %
These dialogue tasks cover generation, classification, and evaluation for both task-oriented and open-ended settings and are drawn from different domains (Figure \ref{fig:tasks}).

Instruction tuned models may ignore instructions and attain good performance with irrelevant prompts~\cite{webson-pavlick-2021}, without actually following user's instructions. 
We address this issue in two ways: (1) we train the models with a variety of outputs given the same input context by creating multiple task formulations, and (2) we propose two instruction-specific meta-tasks (e.g., select an instruction that matches with an input-output pair) to encourage models to adhere to the instructions.

The main contributions of this work are: 
\begin{itemize}[noitemsep,nolistsep]
\item We introduce \dataname, a framework to systematically investigate instruction tuning for dialogue on a large collection of dialogue datasets (59 datasets) and tasks (48 tasks). Our framework is open-sourced and allows easy incorporation and configuration of new datasets and tasks.
\item We show that instruction tuning models enhance zero-shot and few-shot performance on a variety of different dialogue tasks.
\item We provide various analyses and establish baseline and upper bound performance for multiple tasks. We also provide  integration of various task-specific dialogue metrics.
\end{itemize}

Our experiments reveal further room for improvement on issues such as sensitivity to instruction wording and task interference. We hope that \dataname will facilitate further progress on instruction tuning for dialogue tasks.

\section{Related Work}
\label{sec:related}
\vspace{-2mm}
\noindent
\textbf{Pre-training and Multi-Task learning in Dialogue}
Large-scale transformer models~\cite{devlin-etal-2019-bert, Radford2019LanguageMA, brown2020language} pre-trained on massive text corpora have brought substantial performance improvements in natural language processing. Similar trends have occurred in the dialogue domain, where  models such as DialoGPT~\cite{zhang-etal-2020-dialogpt}, Blenderbot~\cite{roller-etal-2021-recipes} and PLATO~\cite{bao-etal-2021-plato} trained on sources such as Reddit or Weibo, or on human-annotated datasets show great capabilities in carrying open-domain conversations. Large-scale pretraining has also shown success in task-oriented dialogue (TOD).  
~\cite{budzianowski-vulic-2019-hello, hosseini2020simple, ham-etal-2020-end,lin-etal-2020-mintl, Yang_Li_Quan_2021} utilized pretrained language models such as GPT-2~\cite{Radford2019LanguageMA} to perform TOD tasks such as language generation or act prediction. Similarly, BERT-type pretrained models have been used for language understanding in TOD tasks~\cite{wu-etal-2020-tod, mi-etal-2021-self}. Several of these works have shown improved performance by performing multi-task learning over multiple tasks~\cite{hosseini2020simple, TACL2889, su2021multitask}. Multi-task pretraining also helps models learn good few-shot capabilities~\cite{wu-etal-2020-tod, peng2021soloist}. Our work covers both open-domain and TOD tasks and goes beyond multi-tasking as it incorporates additional structure of the tasks such as task definitions and constraints.

\noindent
\textbf{Instruction Tuning}
Constructing natural language prompts to perform NLP tasks is an active area of research~\cite{schick-schutze-2021-shot, liu2021pre}. However, prompts are generally short and 
do not generalize well to reformulations and new tasks.
Instruction tuning is a paradigm where models are trained on a variety of tasks with natural language instructions. Going beyond multi-task training, these approaches show better generalization to unseen tasks when prompted with a few examples \cite{bragg2021flex,min2022metaicl,min2022rethinking} or language definitions and constraints
\cite{weller-etal-2020-learning,zhong-etal-2021-adapting-language,xu2022zeroprompt}.
PromptSource~\cite{sanh:iclr22}, FLAN~\cite{flan} and NATURAL INSTRUCTIONS~\cite{mishra2021natural,wangmishra2022benchmarkinggeneralization} collected instructions and datasets for a variety of general NLP tasks. GPT3-Instruct model~\cite{ouyang2022training}
is tuned on a dataset of rankings of model outputs and was trained using human feedback, but it is expensive to train and test.
Instead, our work is tailored to dialogue tasks and incorporates numerous dialogue datasets, tasks, and benchmarks. We show that models trained on collections such as PromptSource are complementary to instruction tuning on dialogue. For dialogue tasks, \citet{madotto2021few} explored prompt-based few-shot learning for dialogue, but without any fine-tuning.
\citet{Mi2021CINSCI} designed task-specific instructions for TOD tasks that improved few-shot performance on several tasks. Our work covers a far greater variety of dialogue domains and datasets in comparison.

\section{Methodology}
\label{sec:method}
In this section, we first discuss instruction tuning setup. Next, we discuss the taxonomy of dialogue tasks, the task meta-information schema, and discuss how dialogue datasets and tasks are mapped into our schema. Finally, we discuss model training and fine-tuning details.

\begin{figure*}[!htb]
\centering
\includegraphics[width=0.9\textwidth]{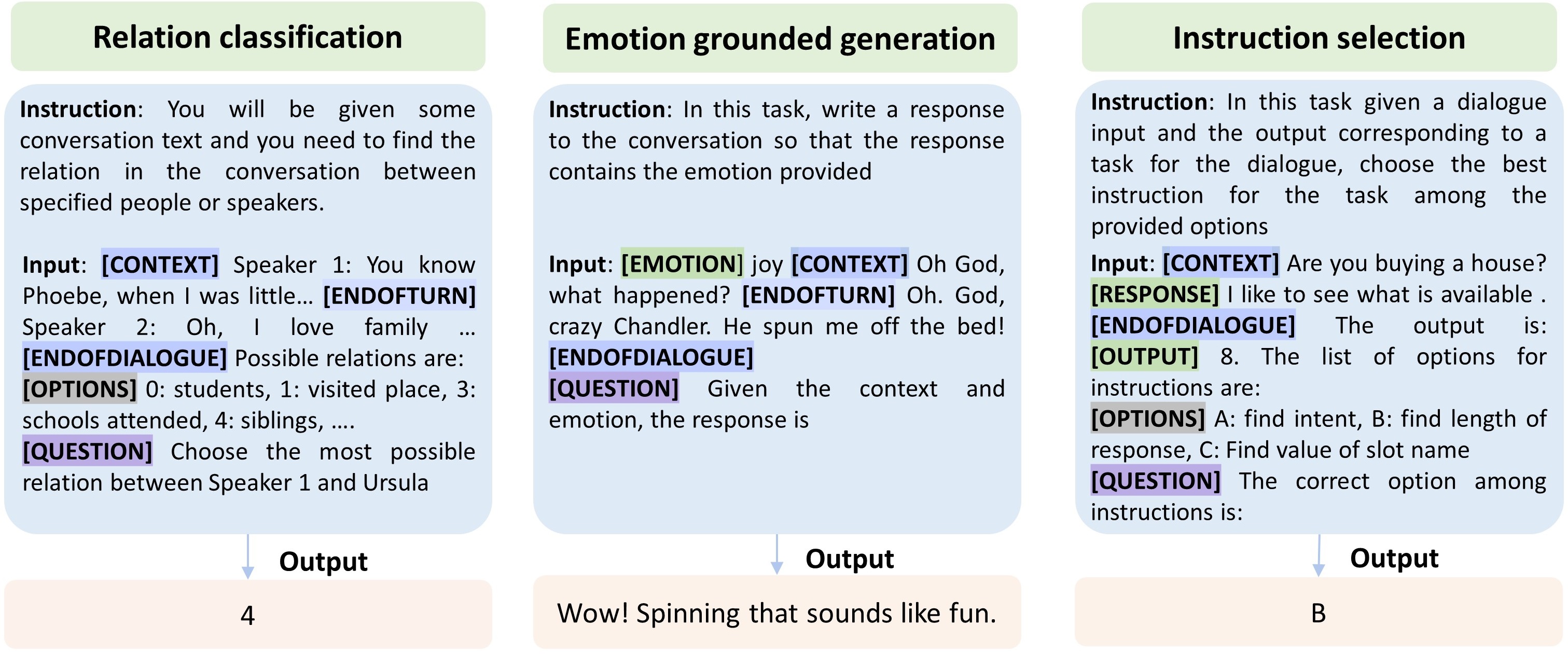}
\caption{Instruction based input-output samples for three tasks. Each task is formatted as a natural language sequence. Each input contains an instruction, instance, optional task-dependent inputs (e.g., class options in relation classification), and task-specific prompts. The instructions and the input instances are formatted using
special tokens such as {[CONTEXT]} and {[QUESTION]}. The Instruction Selection task is a meta-task described in Section~\ref{sec:metatasks}}
\label{fig:taskformat}
\end{figure*}

\subsection{Instruction Tuning Background}
\label{sec:background}

A supervised setup for a dialogue task \textit{t} consists of training instances $d^{t}_{train}\ni(x_i,y_i)$, where $x_i$ and $y_i$  are an input-output pair. A model ${M}$ is trained on $d^{t}_{train}$ and tested on $d_{test}^{t}$. In a cross-task setup, the model $M$ is tested on test instances $d_{test}^{\hat{t}}$ of an unseen task $\hat{t}$. %
In instruction tuning, the model $M$ is provided additional signal or meta information about the task. The meta information can consist of prompts, task definitions, constraints, and examples, and guides the model $M$ towards the expected output space of the unseen task $\hat{t}$.

\subsection{Task Collection} \label{sec:tasks}
We adopt the definition of a task from \citet{sanh:iclr22}, which defined a task as "a general NLP ability that is tested by a group of specific datasets". In \dataname, each task is created from one or more existing open-access dialogue datasets.
Figure~\ref{fig:tasks} shows the taxonomy of dialogue tasks in \dataname, and Table~\ref{tab:tasksdatasets} shows the list of datasets used in each task.
In our taxonomy, \textit{Classification tasks} consist of tasks such as intent classification with a set of predefined output classes. \textit{Generation tasks} consist of tasks such as open-domain, task-oriented, controlled, and grounded response generation, and summarization. 
\textit{Evaluation tasks} consist of response selection in addition to relevance and rating prediction tasks.
\textit{Edit tasks} involve editing a corrupted dialogue response into a coherent response. Corrupted responses are created through shuffling, repeating, adding, or removing phrases/sentences in the gold response.
\textit{Pretraining tasks} involve tasks such as infilling or finding the index of an incoherent or missing utterance. They include multiple tasks covered in prior pretraining work~\cite{mehri-etal-2019-pretraining, zhao-etal-2020-learning-simple, Whang_Lee_Oh_Lee_Han_Lee_Lee_2021, Xu_Tao_Jiang_Zhao_Zhao_Yan_2021}.
\textit{Safety Tasks} consist of toxicity detection, non-toxic, and recovery response generation.
\textit{Miscellaneous tasks} are a set of tasks that belong to specialized domains such as giving advice or persuading a user. 

\subsection{Task Schema and Formatting}
\label{sec:formatting}
All tasks in \dataname are expressed in a natural language sequence-to-sequence format. 
Every task instance is formatted with the following properties:
\textit{Task Definition}: Description of the task containing information about how to produce an output given an input.
\textit{Instance Inputs}: Instances from a dataset converted into a sequence.
\textit{Constraints}: Additional metadata or constraints for a task (emotion tag for emotion-based generation, classes for classification).
\textit{Prompt}: Text sequence that connects the instance back to the instruction, expressed as a command or a question.
\textit{Output}: Output of an instance converted into a sequence.

Figure \ref{fig:taskformat} shows examples of instances from 3 tasks. For each task, we manually compose 3-10 task definitions and prompts. For every instance, a task definition and a prompt are selected randomly during test. We do not include in-context examples in the task schema since dialogue contexts are often long and concatenating long examples would exceed the maximum allowable input length for most models.
Input instances are formatted using special tokens. The token \texttt{[CONTEXT]} signals the start of dialogue content. Dialogue turns are separated by \texttt{[ENDOFTURN]}. \texttt{[ENDOFDIALOGUE]} marks the end of the dialogue and \texttt{[QUESTION]} marks the start of the prompt text. We also incorporate task specific special tokens (such as \texttt{[EMOTION]} for emotion classification task). 
We hypothesize that using a consistent structure and formatting across tasks should help the model adopt the structure and novel input fields for unseen tasks better. 

\noindent
\textbf{Classification Options}:
In classification tasks, the model is trained to predict an output that belongs to one of several classes. To make the model aware of output classes available for an unseen task, we append a list of classes from which the model should choose. We adopt the following two formats for representing the classes: (1) \textit{Name list}: list the class names separated by a class separator token such as a comma, and (2) \textit{Indexed list}: list the classes indexed by either alphabets or numbers (such as 1: class A, 2: class B,...) where the model outputs the index corresponding to the predicted class. This representation is useful when the classification options are long in length, such as in the case of response ranking where the model has to output the best response among the provided candidates.

\noindent
\textbf{Custom inputs}:
Some tasks consist of input fields that are unique to the task. For example, emotion grounded generation consists of emotion labels that the model uses for response generation. We append such inputs to the beginning of the instance sequence along with the field label. For example, we pre-pend ``\texttt{[EMOTION]} happy'' to the dialogue context in the emotion generation task.

In Table \ref{tab:prompttasks} in the Appendix we present the list of tasks with sample inputs for each task.

\subsection{Meta Tasks}
\label{sec:metatasks}
A model can learn to perform well on tasks during training by inferring the domain and characteristics of the dataset instead of paying attention to the instructions, and then fail to generalize to new instructions at the test time. We introduce two meta-tasks that help the model learn the association between the instruction, the data, and the task.
In the \textit{Instruction selection task}, the model is asked to select the instruction which corresponds to a given input-output pair. In the \textit{Instruction binary task}, the model is asked to predict ``yes'' or ``no'' if the provided instruction leads to a given output from an input. We show an example for instruction selection task in Figure \ref{fig:taskformat}.

\subsection{None-of-the-above Options}
\label{sec:nota}
For classification tasks, most tasks assume that the ground truth is always present in the candidate set, which is not the case for all unseen tasks. To solve this issue, we propose adding a NOTA (`None of the above'') option in the classification tasks during training as both correct answers and distractors following~\citet{feng-etal-2020-none} for 10\% of the training instances. 
To add NOTA as a correct answer, we add ``none of the above'' as a classification label option, remove the gold label from the options and set the output label as NOTA. To add NOTA as a distractor, we add NOTA to the classification labels list but keep the gold label as the output label.

\section{Experimental Setup}
\label{sec:setup}

\subsection{Model Details}
\label{sec:model}
Our models use an encoder-decoder architecture and are trained using maximum likelihood training objective. We finetune the following two base models on the tasks from \dataname: 
\begin{enumerate}[noitemsep,nolistsep,leftmargin=*]
  \item T0-3B~\cite{sanh:iclr22} a model initialized from the 3B parameters version of T5~\cite{lester-etal-2021-power}. T0-3B is trained on a multitask mixture of general non-dialogue tasks such as question answering, sentiment detection, and paraphrase identification.
  \item BART0~\cite{Lin2022UnsupervisedCG}, a model with 406 million parameters (8x smaller than T0-3B) based on Bart-large~\cite{lewis-etal-2020-bart}, trained on the same task mixture as T0-3B.
\end{enumerate}
We name the BART0 model tuned on \dataname as \textbf{\modelnamebartzero} and T0-3B model tuned on \dataname as \textbf{\modelnametzero}.
\modelnamebartzero is our main model for experiments since its base BART0 has shown comparable zero-shot performance to T0~\cite{Lin2022UnsupervisedCG} despite being 8 times smaller, whereas the 3B parameter model \modelnametzero is large and impractical to use on popular affordable GPUs.
We perform finetuning on these two models since they both are instruction-tuned on general NLP tasks and thus provide a good base for building a dialogue instruction tuned model.

\subsection{Training Details}
\label{sec:trainingdetails}

For training data creation, we first generate instances from all datasets belonging to each task. We then sample a fixed maximum of $N=5000$ instances per task. Each instance in a task is assigned a random task definition and prompt. We truncate the input sequences to 1024 tokens and target output sequences to 256 tokens. We train \modelnamebartzero on 2 Nvidia 2080Ti GPUs using a batch size of 2 per GPU with gradient checkpointing. We train \modelnametzero on 2 Nvidia A6000 GPUs using a batch size of 1 per GPU with gradient checkpointing. Additional implementation details are present in Appendix~\ref{sec:extraimplementation}.

\begin{table*}[!tb]
\centering
\small
\begin{tabular}{lrrrrrrrrrrrr}
\toprule
Model & \multicolumn{1}{c}{ES} & \multicolumn{1}{c}{AS} & \multicolumn{1}{c}{RC} & \multicolumn{1}{c}{DC} & \multicolumn{4}{c}{BW} & \multicolumn{4}{c}{KG} \\
\cmidrule(lr){2-2}\cmidrule(lr){3-3}\cmidrule(lr){4-4}\cmidrule(lr){5-5}
\cmidrule(lr){6-9} \cmidrule(lr){10-13}
  & {ACC} & {ACC} & {ACC} & {ACC} & {ACC} & {B-2} & {R-L} & \multicolumn{1}{c}{GR} & F1 & {B-2} & {R-L} & \multicolumn{1}{c}{GR} \\ \hline 
 \rowcolor{Gray}
\multicolumn{13}{l}{\textit{Baselines and Our Models}}
 \\
BART0 & 22.2 & 58.5 & 6.3 & 33.7 & 4.2 & 4.9 & 12.0 & 45.7 & 17.4 & 5.3 & 13.3 & 23.9 \\
T0-3B & 45.9 & 60.2 & 1.3 & 33.1 & 14.1 & 4.1 & 10.7 & 55.5 & 14.2 & 3.2 & 10.7 & 78.0 \\
GPT-3 & 57.5 & 56.5 & 11.5 & 37.3 & 16.5 & 7.2 & 15.7 & 57.0 & 18.5 & 3.9 & 11.6 & \textbf{83.8} \\
\modelnamebartzero (Ours) & 66.7 & 59.5 & \textbf{17.8} & \textbf{35.6} & \textbf{56.3} & \textbf{13.1} & 26.4 & 60.2 & \textbf{27.8} & \textbf{11.1} & \textbf{21.4} & 68.5 \\
\modelnametzero (Ours) & \textbf{74.4} & \textbf{65.2} & 6.4 & 34.5 & 55.0 & 12.4 & \textbf{26.5} & \textbf{61.3} & 22.2 & 7.2 & 16.5 & 69.8 \\\hline
 \rowcolor{Gray}
\multicolumn{13}{l}{\textit{Few and Full shot Variations}}
 \\
DB-Few & 77.1 & 69.1 & 28.0 & 43.0 & 72.2 & 16.7 & 30.7 & 60.3 & 27.9 & 9.7 & 20.0 & 68.0 \\
DB-Full & 90.7 & 83.3 & 62.7 & 77.4 & 83.7 & 20.8 & 33.8 & 61.0 & 30.9 & 11.6 & 22.8 & 70.5 \\
\hline
 \rowcolor{Gray}
\multicolumn{13}{l}{\textit{Model Ablations for \modelnamebartzero}}
 \\
DB-no-base & 40.1 & 52.7 & {17.1} & 35.1 & 53.9 & 12.0 & 26.6 & 57.8 & 29.8 & 12.0 & 22.8 & 69.6 \\
DB-no-instr & 23.0 & 43.2 & 15.1 & 35.4 & 50.0 & 13.0 & 27.0 & 61.1 & 30.1 & 11.2 & 20.8 & 65.7 \\
DB-no-nota & 66.5 & 57.2 & 17.2 & 35.9 & 56.1 & 10.9 & 25.3 & 58.4 & 28.0 & 11.0 & 21.4 & 67.6 \\
DB-no-meta & 44.5 & 52.0 & 14.1 & 35.4 & 52.5 & 14.1 & 28.1 & 61.3 & 29.6 & 11.8 & 22.1 & 70.5\\
\specialrule{.8pt}{0pt}{2pt}
\end{tabular}
\caption{Zero-shot evaluation on unseen tasks. B-2 stands for BLEU2, R-L for RougeL and GR for GRADE metric. Here ES stands for Eval Selection, AS for Answer Selection, RC for Relation Classification, DC for Dialfact Classification, BW for Begins With, KG for Knowledge Grounded generation. DB-Few and DB-Full are variants of \modelnamebartzero. Our models \modelnamebartzero and \modelnametzero outperform the baseline models and their ablated versions.}
\label{tab:mainresults}
\vspace{-2mm}
\end{table*}

\section{Experiments and Results}
\label{sec:results}
We evaluate our models on multiple zero-shot and few-shot settings. We establish benchmark results for Zero-shot unseen tasks evaluation (Section \ref{sec:zeroshot}) and Response evaluation task (Section \ref{sec:zeroeval}) and perform error analysis. Next, we perform zero-shot and few-shot experiments on three important dialogue tasks: intent detection, slot value generation, and dialogue state tracking (Section \ref{sec:dialogtasks}).

\subsection{Zero-shot Unseen Tasks Evaluation}
\label{sec:zeroshot}
In this experiment, we test our models' zero-shot ability on tasks not seen during training. 

\subsubsection{Unseen Tasks for Zero-shot Setting}
\label{sec:zeroshottasks}
We perform evaluation on the test set of the following 6 tasks not seen during training:
\begin{enumerate}[noitemsep,nolistsep,leftmargin=*]
  \item \textit{Dialfact classification}: predict if an evidence supports, refutes, or does not have enough information to validate the response
  \item \textit{Relation classification}: predict the relation between two people in a dialogue
  \item \textit{Answer selection}: predict an answer to a conversational question
  \item \textit{Eval selection}: choose the most relevant response among the provided 4 options. Dataset and ratings based on DSTC 10 Automatic evaluation challenge~\cite{chen2021automatic}
  \item \textit{Knowledge grounded generation}: generate a response based on background knowledge
  \item \textit{Begins with generation}: generate a response that starts with the provided initial phrase
\end{enumerate}

All 6 tasks have varying levels of difficulty and cover both classification and generation. To emulate a zero-shot scenario, we remove all relation-based, evaluation type, answer generation, and wiki-based tasks from the training task set. 
The set of tasks used for training is presented in Table \ref{tab:datasets}.
We evaluate on the full test sets for Dialfact, relation, and answer classification, and sample 1000 instances for the rest of the tasks.

\subsubsection{Setup and Baselines}
We perform inference and evaluation on the 6 unseen tasks described in Section~\ref{sec:zeroshottasks}. We compare the following models and baselines:

\begin{itemize}[noitemsep,nolistsep,leftmargin=*]
\item {BART0} and {T0-3B} - Models that form a base for our models, trained on a mixture of non-dialogue general NLP tasks (described in Section \ref{sec:model}). 
\item GPT-3~\cite{brown2020language} - Davinci version of GPT-3 tested using our instruction set. 
\item \modelnamebartzero and \modelnametzero  \xspace  - Our models described in Section \ref{sec:model}.
\item DB-Few - Few-shot version of \modelnamebartzero. 100 random training set instances of the test tasks are mixed with the instances of train tasks.
\item DB-Full - Version of \modelnamebartzero where 5000 instances per test tasks are mixed with the instances of the train tasks. This baseline serves as the upper bound for our models' performance.
\end{itemize}

We also experiment with the following ablations of \modelnamebartzero:
\begin{itemize}[noitemsep,nolistsep,leftmargin=*]
\item {DB-no-base} - Uses Bart-large instead of using the BART0 as the base model. 
\item {DB-no-instr} - Trained with no instructions or prompts. Task constraints and class options are still specified. We specify the task name instead of instructions to help the model identify the task.
\item {DB-no-nota} - Trained without None-of-the-above from Section \ref{sec:nota}
\item {DB-no-meta} - Trained without the meta tasks from Section \ref{sec:metatasks}
\end{itemize}

\subsubsection{Results and Discussion}
We present the results for zero-shot experiments in Table~\ref{tab:mainresults} and report the accuracy metric for the Eval selection, Answer selection, Dialfact classification and Relation classification tasks. For Begins with task, we report BLEU2, ROUGEL, and accuracy defined as the proportion of responses that begins with the initial phrase provided. For Knowledge grounded generation we report BLEU2, and ROUGEL metrics along with F1 as defined in \cite{dinan2018wizard}. For the generation tasks we also report the automatic metric GRADE~\cite{huang-etal-2020-grade} (which has shown good correlation with human ratings on response coherence). 
For GPT-3 baseline we report the metrics on 200 randomly sampled instances per task. We average scores obtained across the instructions and prompts.
We notice the following general trends in our results.

\noindent
\textbf{Instruction tuning on \dataname improves performance on unseen dialogue tasks}: The \modelnamebartzero and \modelnametzero models instruction tuned on \dataname achieve better performance on all tasks compared to their base models BART0 and T0-3B. Notably, for the Eval selection, Relation classification and Begins with generation tasks, our models perform about 3 times better than the base models. Our model also performs significantly better than GPT-3 for all tasks except for Dialfact classification.
In the case of the Answer selection task, the difference in performance is lower compared to other models since the baseline models are also trained on similar extractive and multi-choice question answering tasks. 
Relation and Dialfact classification are hard tasks for all models since there are no similar train tasks.

\noindent
\textbf{Larger models are not necessarily better across tasks}: Experiments across varying model size show that while T0-3B and \modelnametzero perform better on the Eval selection and Answer Selection tasks and perform equivalently on the Begins with generation task, BART0 and \modelnamebartzero perform better on the rest of the unseen tasks. While \modelnametzero is better at classification tasks, it has poor performance on generation compared to \modelnamebartzero. We also observed that \modelnametzero sometimes produces empty or repetitive outputs for generation tasks.

\begin{table*}[]
\small
\centering
\setlength\tabcolsep{1.3pt}
\begin{tabular}{lrrrrrrrrrrrrrrr}
    \hline
    Model & DSTC6 & DSTC7 & HUMOD  & TU & PZ & DZ & CG & PU & DGU & DGR & FT & EG & FD & Average \\
    \hline
MAUDE~\shortcite{sinha-etal-2020-learning} & 0.115 & 0.045 & 0.112 & 0.136 & 0.360 & 0.120 & 0.304 & 0.306 & 0.192 & -0.073 & -0.11 & -0.057 & -0.285 & 0.090 \\
GRADE~\shortcite{huang-etal-2020-grade} & 0.121 & 0.332 & \textbf{0.612}& 0.176 & 0.583 & 0.532 & 0.571 & 0.329 & 0.596 & 0.254 & 0.048 & 0.300 & 0.106 & 0.351 \\
USR~\shortcite{mehri-eskenazi-2020-usr} & 0.166 & 0.249 & 0.34  & 0.291 & 0.496 & 0.363 & 0.487 & 0.140 & 0.353 & 0.066 & 0.055 & 0.268 & 0.084 & 0.258 \\
FED~\shortcite{mehri-eskenazi-2020-unsupervised} & -0.082 & -0.070 & -0.077 & -0.090 & -0.232 & -0.080 & -0.137 & -0.004 & 0.025 & -0.009 & 0.173 & 0.005 & 0.178 & -0.031 \\
FlowScore~\shortcite{li-etal-2021-conversations}& 0.095 & 0.067 & -0.049 & 0.068 & 0.202 & -0.063 & - & 0.053 & 0.053 & - & -0.043 & - & -0.009 & 0.029 \\
USL-H~\shortcite{phy-etal-2020-deconstruct}& 0.180 & 0.261 & 0.53 & 0.319 & 0.409 & 0.385 & 0.452 & \textbf{0.493} & 0.481 & 0.09 & 0.115 & 0.237 & 0.202 & 0.320 \\
QuestEval~\shortcite{scialom-etal-2021-questeval} & 0.089 & 0.222 & 0.217& 0.104 & 0.32 & 0.22 & 0.344 & 0.106 & 0.243 & -0.026 & 0.168 & 0.195 & 0.114 & 0.178 \\
DEB~\shortcite{sai-etal-2020-improving}  & 0.214 & 0.351 & 0.649 & 0.123 & 0.579 & 0.486 & 0.504 & 0.351 & 0.579 & \textbf{0.363} & 0.044 & 0.395 & 0.141 & 0.367 \\
DynaEval~\shortcite{zhang-etal-2021-DynaEval}  & 0.252 & 0.066 & 0.112 & -0.013 & 0.165 & 0.169 & 0.202 & 0.148 & 0.038 & 0.122 & 0.247 & 0.159 & \textbf{0.555} & 0.171 \\
DialogRPT~\shortcite{gao-etal-2020-dialogue} & 0.162 & 0.255 & 0.198 & 0.118 & 0.114 & 0.067 & 0.158 & -0.036 & 0.075 & 0.037 & -0.249 & 0.203 & -0.134 & 0.074 \\
\hline
Ours (\modelnametzero) & \textbf{0.553} & \textbf{0.451} & \underline{0.582} & \textbf{0.446} & \textbf{0.651} & \textbf{0.601} & \textbf{0.498} & \underline{0.376} & \textbf{0.634} & \underline{0.286} & \textbf{0.263} & \textbf{0.475} & \underline{0.228} & \textbf{0.465} \\ 
\hline
    \end{tabular}
    \caption{Spearman correlation of model predictions with human ratings. Bold and underlined scores represent the evaluation sets on which our model performs the best and second best respectively. We also present the macro average scores.
    TU, PU, PZ, DZ, CG, DGU, DGR, EG, FT and FD are abbreviations for
    TopicalChat-USR, PersonaChat-USR~~\cite{mehri-eskenazi-2020-usr},
    PersonaChat-Zhao~\cite{zhao-etal-2020-designing},
    DailyDialog-Zhao~\cite{zhao-etal-2020-designing}, ConvAI2-GRADE~\cite{huang-etal-2020-grade}, DailyDialog-Gupta~\cite{gupta-etal-2019-investigating}, DailyDialog-GRADE~\cite{huang-etal-2020-grade}, Empathetic-GRADE~\cite{huang-etal-2020-grade}, FED-Turn and FED-Dial~\cite{mehri-eskenazi-2020-unsupervised}. \modelnametzero is ranked the first or second best in the majority of the evaluation sets.}
    \label{tab:eval}
    \vspace{-2mm}
\end{table*}

\begin{figure}[t]
\centering
\includegraphics[width=0.42\textwidth]{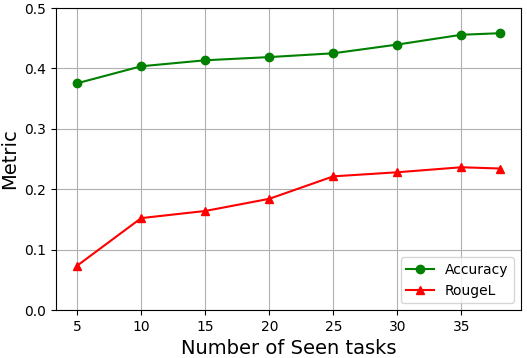}
\caption{
Model's performance on unseen tasks improves with the number of seen tasks during training. We report average Accuracy across Eval Selection, Answer Selection, Relation Classification, and Dialfact Classification, and average RougeL scores for Knowledge Grounded Generation and Begins with Generation.
}
\label{fig:numtasks}
\vspace{-1em}
\end{figure}

\noindent
\textbf{Few-shot training significantly improves performance}:
DB-Few model that incorporates 100 instances per test task in its training data shows significant improvements in performance compared to its zero-shot counterpart \modelnamebartzero. We see about 12-16\% improvements on the Eval selection, Answer selection, and Dialfact classification tasks, and 30-50\% improvement on the Begins with and Relation classification tasks.

\noindent
\textbf{Full-shot training can improve performance across multiple tasks}:
DB-Full model achieves high performance across all test tasks. The full-shot performance of \modelnamebartzero on Dialfact and relation classification tasks are near state-of-the-art performance without using the full train datasets.

\noindent
\textbf{Meta tasks and NOTA are important for better generalization}:
We see a large performance drop on unseen classification tasks when meta tasks (see Section \ref{sec:metatasks}) are removed. This shows that meta tasks help the model develop better representations and understanding of natural language instructions. DB-no-nota shows a slight performance drop in the classification task, indicating NOTA objective is helpful, but not crucial for performance.

\noindent
\textbf{Pretraining on general NLP tasks helps dialogue instruction tuning}:
DB-no-base model shows a high performance drop on Eval selection and Answer selection tasks, and a small drop on other test tasks. We conclude that instruction tuning for general NLP tasks helps dialogue instruction tuning.

\noindent
\textbf{Using instructions leads to better generalization}:
DB-no-instr shows worse performance than \modelnamebartzero on all tasks, especially on Eval selection, Answer selection, and Relation classification tasks. This indicates that training with instructions is crucial for zero-shot performance on unseen tasks.

\noindent
\textbf{Training on more seen tasks improves generalization on unseen tasks}:
In Figure \ref{fig:numtasks} we show the impact of varying the number of seen tasks on the performance on unseen tasks. We adopt the train-test task split from section \ref{sec:zeroshot}. We observe that the performance improves sharply up to 20-25 tasks and then further keeps steadily increasing with each new task. This indicates that increasing the number of tasks can lead to better zero-shot generalization and that scaling to more tasks may lead to better instruction-tuned models.

\subsubsection{Analysis}
\label{sec:erroranalysis}
\noindent
\textbf{Sensitivity to instruction wording}: To analyze the sensitivity of our models to instruction wording, we breakdown the evaluation metrics per unique instruction used during inference for the \modelnamebartzero model. The accuracy varies from 65.6-67.8 across instructions for Eval selection, from 52.5 to 75.0 for Answer selection, 17.1 to 18.4 for Relation classification, 34.7 to 37.1 for Dialfact classification, 49.8 to 62.3 for Begins with generation, and F1 score varies from 26.6 to 28.6 for Knowledge grounded generation. 
Thus, our model is moderately sensitive to the instruction wording. 

\noindent
\textbf{Errors in model outputs}:
We perform qualitative analysis of randomly sampled outputs of the models. For classification tasks, a common error across all models is generating outputs outside of the provided list of classes. This happens with GPT-3 for $20\%$, BART0 $10\%$ and T0-3B $17.8\%$ of the inputs, but for \modelnamebartzero and \modelnametzero this occurs only for $2.5\%$ and $4.8\%$ of the inputs. Other possible but rare types of errors include copying the provided input as output, early truncation of generated responses, and performing an unspecified task.
Apart from the unseen task set adopted for our experiments in section \ref{sec:zeroshottasks}, we tried other seen-unseen task configurations and found that both our models and baselines models cannot perform certain tasks such as Infilling missing utterance, Recovery response generation, and Ends with response generation in a zero-shot manner. However, the models could quickly learn these tasks when trained on a few task instances.

\textit{In Table~\ref{tab:exampleoutputs} of Appendix~\ref{sec:sampleconversation} we provide a sample conversation}, various instructions for that conversation, and the outputs generated by \modelnamebartzero based on the specified instructions.

\subsection{Zero-shot Automatic Response Evaluation}
\label{sec:zeroeval}
Development of automatic dialogue metrics that show high correlations with human judgements is a challenging and crucial task for dialogue systems. 
Automated metrics such as BLEU \cite{papineni2002bleu} and METEOR \cite{banerjee2005meteor} correlate poorly with human judgement \cite{gupta-etal-2019-investigating}.
In this experiment, we test our model's  zero-shot automatic evaluation capabilities through the Eval Relevance task.
We use the evaluation ratings released in the DSTC-10 Automatic evaluation challenge~\cite{chen2021automatic} that consists of 65,938 context-response pairs along with corresponding human ratings aggregated across various evaluation sets.
We train a version of \modelnametzero on tasks excluding any eval tasks (shown in Table \ref{tab:datasets}).
Given a dialogue context and a candidate response, we instruct the model to predict ``yes'' if the response is relevant to the context, otherwise predict ``no''. 
We calculate the probability of ``yes'' as $p(yes) = p(yes)/(p(yes)+p(no))$.
We calculate the Spearman correlation of the model's prediction with human ratings for relevance provided in the DSTC-10 test sets, and present the results in Table~\ref{tab:eval}.
We compare our model with reference-free models studied in \citet{yeh-etal-2021-comprehensive}. 
\modelnametzero is ranked the first or second in the majority of the evaluation datasets. Our model learns coherence from the variety of tasks it is trained on and demonstrates high zero-shot dialogue evaluation capabilities.

\subsection{Zero-shot and Few-shot Dialogue Tasks}
\label{sec:dialogtasks}
We test the zero-shot and few-shot abilities of our models on three important dialogue tasks: intent prediction, slot filling, and dialogue state tracking.

\begin{table}[!tb]
\centering
\small
\begin{tabular}{lr}

    \hline
    Model & Accuracy \\
    \hline
    ConvERT \citep{casanueva-etal-2020-efficient} & 83.32 \\
    ConvERT + USE \citep{casanueva-etal-2020-efficient} & 85.19 \\
    Example-Driven \citep{mehri-eric-2021-example}  & 85.95  \\ 
    PPTOD$_{base}$~\cite{su-etal-2022-multi} & 82.81 \\
    PPTOD$_{large}$~\cite{su-etal-2022-multi} & 84.12 \\
    \modelnamebartzero (Ours) & 84.30 \\
    \hline
    BART0 (zero-shot) & 14.72\\
    \modelnamebartzero (Ours, zero-shot) & 58.02 \\
    \hline
    \end{tabular}
    \caption{Intent prediction accuracy on the \textsc{Banking77} corpus \citep{casanueva-etal-2020-efficient}. Models in the first section of the table are trained in a few-shot setting with 10 instances per intent. Models in the second section are tested in a zero-shot setting.}
    \label{tab:intent}
    \vspace{-2mm}
\end{table}

\subsubsection{Intent Prediction} Intent prediction is the task of predicting an intent class for a given utterance. 
We conduct few-shot experiments on the Banking77 benchmark dataset~\cite{casanueva-etal-2020-efficient} that
contains 77 unique intent classes. Models are trained on 10 instances per test intent class. We compare our model \modelnamebartzero with Convert Models \citep{casanueva-etal-2020-efficient} that are Bert-based dual encoder discriminative models and PPTOD~\cite{su-etal-2022-multi}, a model pre-trained on multiple task-oriented dialogue datasets. For this experiment, \modelnamebartzero is pretrained on the training task mixture from Section~\ref{sec:zeroshottasks} that includes few intent detection datasets except for Banking77 dataset.
The results in Table \ref{tab:intent} shows that our model is able to attain competitive performance in the few-shot setting, without necessitating complex task-specific architectures or training methodology. It is notable that \modelnamebartzero performs better than PPTOD which uses about about two times more parameters and is trained similarly to our model using a Seq2Seq format. We also note that while BART0 model struggles in zero-shot setting, \modelnamebartzero shows greatly improved performance.

\begin{table}[t]
\centering
\small
\begin{tabular}{lr}

    \hline
    Model & \multicolumn{1}{c}{F1} \\
    \hline
    \textsc{ConVEx {\small \citep{henderson2020convex}}}  &  \multicolumn{1}{c}{5.2} \\

\textsc{Coach+TR {\small \citep{liu2020coach}}}  &  \multicolumn{1}{r}{10.7} \\

\textsc{GenSF {\small \citep{mehri2021gensf}}}  &  \multicolumn{1}{r}{19.5} \\

\modelnamebartzero (Ours)  &  \multicolumn{1}{r}{{56.4}} \\

    \hline
    \end{tabular}
    \caption{Zero-shot slot filling results on the Restaurant8k corpus.}
    \label{tab:slotzeroshot}
        \vspace{-3mm}
\end{table}

\begin{table}[t]
\centering
\small
\begin{tabular}{lrr}
    \hline
    Domain & \multicolumn{1}{c}{\textsc{GenSF}} & \multicolumn{1}{c}{\modelnamebartzero (Ours)}\\
    \hline
    Buses & 90.5 & 97.8\\
    Events & 91.2 & 94.3\\
    Homes & 93.7 & 96.5 \\
    Rental Cars & 86.7 & 94.2 \\

    \hline
    \end{tabular}
    \caption{Few-shot slot filling F1 scores on DSTC8 data.}
    \label{tab:slotfewshot}
    \vspace{-2em}
\end{table}

\subsubsection{Slot Filling}
Slot filling is the problem of detecting slot values in a given utterance. We carry out zero-shot experiments on the Restaurant8k corpus \citep{coope2020span} and few-shot experiments on the DSTC8 dataset \cite{rastogi2020schema}, demonstrating significant performance gains over prior work. In the zero-shot experiments, the training set includes several slot filling datasets except for the Restaurant8k dataset used for testing. 
Table \ref{tab:slotzeroshot} shows that our approach attains a 36.9 point improvement in zero-shot slot filling. This result especially highlights the efficacy of instruction tuning at leveraging large-scale pretrained language models to generalize to unseen tasks. 
The few-shot slot filling experiments on the DSTC8 datasets span four domains - buses, events, homes, rental cars and involves training on $25\%$ of the training dataset. The set of tasks used for training the model are presented in Table~\ref{tab:datasets}.
We see significant improvement compared to the baseline in the few-shot setting on the DSTC8 benchmark in Table \ref{tab:slotfewshot}.

\begin{table}[t]
\small
\centering
\begin{tabular}{lrr}
    \hline
    Model & 1\% data & 5\% data \\
    \hline
    PPTOD$_{base}$ & 29.7 & 40.2 \\
    \modelnamebartzero (Ours) & 29.2 & 38.1 \\
    \hline
    \end{tabular}
    \caption{Joint goal accuracy for dialogue state tracking in few-shot setting on  1\% and 5\% data of Multiwoz.}
    \label{tab:dst}
    \vspace{-1em}
\end{table}

\subsubsection{Dialogue State Tracking}
We  evaluate our model on the dialogue state tracking task which involves filling in values of pre-defined slots. We adopt the experimental setup from PPTOD \cite{su2021multitask}, and conduct few-shot experiments on MultiWOZ 2.0 \cite{budzianowski-etal-2018-multiwoz}. Similar to PPTOD, our \modelnamebartzero model is first pre-trained on 7 datasets: KVRET \cite{eric-etal-2017-key}, WOZ \cite{mrksic-etal-2017-neural}, CamRest676 \cite{wen-etal-2017-network}, MSR-E2E \cite{li2018microsoft}, Frames \cite{el-asri-etal-2017-frames}, TaskMaster \cite{byrne-etal-2019-taskmaster}, Schema-Guided \cite{rastogi2020towards} along with other non-related dialogue tasks. We then train on 1\% and 5\% splits of MultiWOZ for 40 epochs with a learning rate of $5e-5$. In Table~\ref{tab:dst} we present  few-shot dialogue state tracking results on the MultiWOZ test set. We find that our model obtains 29.2 and 38.1 joint goal accuracy on the 1\% and 5\% training data splits, respectively. Our results demonstrate that our model performs well on few-shot dialogue state tracking, and achieves competitive results against PPTOD which is twice the size of our model.

\section{Conclusion}
We propose \dataname, an instruction tuning
framework for dialogue, which contains multiple dialogue tasks created from 
openly available dialogue datasets. 
We also propose two meta-tasks to encourage the model to pay attention to instructions.
Our results show that models trained on \dataname achieve good zero-shot performance on unseen tasks (e.g., dialogue evaluation) and good few-shot performance on dialogue tasks (e.g., intent prediction, slot filling).  We perform ablation studies showing the impact of using an instruction tuned base model, model size/type, increasing the number of tasks, and incorporating our proposed meta tasks.
Our experiments reveal that instruction tuning does not benefit all unseen test tasks and that improvements can be made in instruction wording invariance and task interference.
We hope that \dataname will facilitate further progress on instruction-tuning systems for dialogue tasks.

\section{Limitations}
Our work is the first to explore instruction tuning for dialogue and establishes baseline performance for a variety of dialogue tasks. However, there is room for improvements in the following aspects: 
1) Unlike a few prior works, the instructions and prompts used in this work are not crowdsourced and are limited in number. Furthermore, our instructions and tasks are only specified in the English language. Future work may look into either crowdsourcing or automatic methods for augmenting the set of instructions in terms of both language diversity as well as quantity.
2) Instruction tuning does not show significant improvements in zero-shot setting on a few tasks such as relation classification and infilling missing utterances in our experiments. Future work can look into investigating why certain tasks are more challenging than others for zero-shot generalization.  Furthermore, zero-shot performance of our models on many tasks is still far from the few-shot and full-shot performance on those tasks. We hope that \dataname can be lead to further investigations and improvements in this area.
3) We observed a few instances of task interference in our experiments. For example, the set of tasks used for zero-shot automatic response evaluation as mentioned in Table~\ref{tab:datasets} is different and smaller from the set of tasks used in our main experiments in Section \ref{sec:zeroshottasks}. We found that incorporating a few additional tasks lead to a reduction in the performance on zero-shot automatic response evaluation. 
Furthermore, training on multiple tasks can lead to task forgetting. 
To address these issues, future work can take inspiration from work related to negative task interference~\cite{wang-etal-2020-negative,larson2022redwood}, transferability~\cite{vu-etal-2020-exploring,wu2020understanding, xing2022balancing} and lifelong learning~\cite{wang-etal-2020-efficient}.
4) Our models are sensitive to the wording of the instructions, especially in zero-shot settings as discussed in Section~\ref{sec:erroranalysis}. Improving insensitivity to prompts and instructions is an important future research direction.
5) Our work does not explore in-context few-shot learning through examples as the prompt length can go beyond models' maximum input length. It also does not study the composition of multiple tasks through instructions. Both these aspects warrant further investigations. 6) \dataname includes only text based tasks, and future work may look into incorporating datasets with other modalities such as vision and audio. 

\section{Ethics and Broader Impact}
\textbf{Broader Impact and applications:} Our framework leverages instruction tuning on multiple dialogue tasks, allowing multiple functionalities to be quickly implemented and evaluated in dialogue systems. For example, tasks pertaining to both task-oriented dialogue tasks, such as slot detection and domain-specific tasks such as emotion detection can be added and evaluated against state-of-the-art dialogue systems. This enables users to diagnose their models on different tasks and expand the abilities of multi-faceted dialogue systems, which can lead to richer user interactions across a wide range of applications. Our framework allows training models below billion parameter range, making them more accessible to the research community.

\noindent\textbf{Potential biases:} Current conversational systems suffer from several limitations, and lack empathy, morality, discretion, and factual correctness.  Biases may exist across datasets used in this work and those biases can propagate during inference into the unseen tasks. Few-shot and zero-shot methods are easier to train, and their use can lead to a further increase of both the benefits and risks of models. To mitigate some of those risks, we have included tasks and datasets in our framework that encourage safety such as ToxiChat for toxic response classification task and SaFeRDialogues for recovery response generation task, and that improve empathy such as EmpatheticDialogues for empathy.

\bibliography{anthology,custom}

\begin{thebibliography}{123}
\expandafter\ifx\csname natexlab\endcsname\relax\def\natexlab#1{#1}\fi

\bibitem[{Baheti et~al.(2021)Baheti, Sap, Ritter, and
  Riedl}]{baheti-etal-2021-just}
Ashutosh Baheti, Maarten Sap, Alan Ritter, and Mark Riedl. 2021.
\newblock \href {https://doi.org/10.18653/v1/2021.emnlp-main.397} {Just say no:
  Analyzing the stance of neural dialogue generation in offensive contexts}.
\newblock In \emph{Proceedings of the 2021 Conference on Empirical Methods in
  Natural Language Processing}, pages 4846--4862, Online and Punta Cana,
  Dominican Republic. Association for Computational Linguistics.

\bibitem[{Banerjee and Lavie(2005)}]{banerjee2005meteor}
Satanjeev Banerjee and Alon Lavie. 2005.
\newblock Meteor: An automatic metric for mt evaluation with improved
  correlation with human judgments.
\newblock In \emph{Proceedings of the acl workshop on intrinsic and extrinsic
  evaluation measures for machine translation and/or summarization}, pages
  65--72.

\bibitem[{Bao et~al.(2021)Bao, He, Wang, Wu, Wang, Wu, Guo, Liu, and
  Xu}]{bao-etal-2021-plato}
Siqi Bao, Huang He, Fan Wang, Hua Wu, Haifeng Wang, Wenquan Wu, Zhen Guo,
  Zhibin Liu, and Xinchao Xu. 2021.
\newblock \href {https://doi.org/10.18653/v1/2021.findings-acl.222} {{PLATO-2}:
  Towards building an open-domain chatbot via curriculum learning}.
\newblock In \emph{Findings of the Association for Computational Linguistics:
  ACL-IJCNLP 2021}, pages 2513--2525, Online. Association for Computational
  Linguistics.

\bibitem[{Bragg et~al.(2021)Bragg, Cohan, Lo, and Beltagy}]{bragg2021flex}
Jonathan Bragg, Arman Cohan, Kyle Lo, and Iz~Beltagy. 2021.
\newblock Flex: Unifying evaluation for few-shot nlp.
\newblock In \emph{Advances in Neural Information Processing Systems (NeurIPS
  2021)}.

\bibitem[{Brown et~al.(2020)Brown, Mann, Ryder, Subbiah, Kaplan, Dhariwal,
  Neelakantan, Shyam, Sastry, Askell et~al.}]{brown2020language}
Tom Brown, Benjamin Mann, Nick Ryder, Melanie Subbiah, Jared~D Kaplan, Prafulla
  Dhariwal, Arvind Neelakantan, Pranav Shyam, Girish Sastry, Amanda Askell,
  et~al. 2020.
\newblock Language models are few-shot learners.
\newblock \emph{Advances in neural information processing systems},
  33:1877--1901.

\bibitem[{Budzianowski and Vuli{\'c}(2019)}]{budzianowski-vulic-2019-hello}
Pawe{\l} Budzianowski and Ivan Vuli{\'c}. 2019.
\newblock \href {https://doi.org/10.18653/v1/D19-5602} {Hello, it{'}s {GPT}-2 -
  how can {I} help you? towards the use of pretrained language models for
  task-oriented dialogue systems}.
\newblock In \emph{Proceedings of the 3rd Workshop on Neural Generation and
  Translation}, pages 15--22, Hong Kong. Association for Computational
  Linguistics.

\bibitem[{Budzianowski et~al.(2018)Budzianowski, Wen, Tseng, Casanueva, Ultes,
  Ramadan, and Ga{\v{s}}i{\'c}}]{budzianowski-etal-2018-multiwoz}
Pawe{\l} Budzianowski, Tsung-Hsien Wen, Bo-Hsiang Tseng, I{\~n}igo Casanueva,
  Stefan Ultes, Osman Ramadan, and Milica Ga{\v{s}}i{\'c}. 2018.
\newblock \href {https://doi.org/10.18653/v1/D18-1547} {{M}ulti{WOZ} - a
  large-scale multi-domain {W}izard-of-{O}z dataset for task-oriented dialogue
  modelling}.
\newblock In \emph{Proceedings of the 2018 Conference on Empirical Methods in
  Natural Language Processing}, pages 5016--5026, Brussels, Belgium.
  Association for Computational Linguistics.

\bibitem[{Byrne et~al.(2019)Byrne, Krishnamoorthi, Sankar, Neelakantan,
  Goodrich, Duckworth, Yavuz, Dubey, Kim, and
  Cedilnik}]{byrne-etal-2019-taskmaster}
Bill Byrne, Karthik Krishnamoorthi, Chinnadhurai Sankar, Arvind Neelakantan,
  Ben Goodrich, Daniel Duckworth, Semih Yavuz, Amit Dubey, Kyu-Young Kim, and
  Andy Cedilnik. 2019.
\newblock \href {https://doi.org/10.18653/v1/D19-1459} {Taskmaster-1: Toward a
  realistic and diverse dialog dataset}.
\newblock In \emph{Proceedings of the 2019 Conference on Empirical Methods in
  Natural Language Processing and the 9th International Joint Conference on
  Natural Language Processing (EMNLP-IJCNLP)}, pages 4516--4525, Hong Kong,
  China. Association for Computational Linguistics.

\bibitem[{Casanueva et~al.(2020)Casanueva, Tem{\v{c}}inas, Gerz, Henderson, and
  Vuli{\'c}}]{casanueva-etal-2020-efficient}
I{\~n}igo Casanueva, Tadas Tem{\v{c}}inas, Daniela Gerz, Matthew Henderson, and
  Ivan Vuli{\'c}. 2020.
\newblock \href {https://doi.org/10.18653/v1/2020.nlp4convai-1.5} {Efficient
  intent detection with dual sentence encoders}.
\newblock In \emph{Proceedings of the 2nd Workshop on Natural Language
  Processing for Conversational AI}, pages 38--45, Online. Association for
  Computational Linguistics.

\bibitem[{Chawla et~al.(2021)Chawla, Ramirez, Clever, Lucas, May, and
  Gratch}]{chawla-etal-2021-casino}
Kushal Chawla, Jaysa Ramirez, Rene Clever, Gale Lucas, Jonathan May, and
  Jonathan Gratch. 2021.
\newblock \href {https://doi.org/10.18653/v1/2021.naacl-main.254}
  {{C}a{S}i{N}o: A corpus of campsite negotiation dialogues for automatic
  negotiation systems}.
\newblock In \emph{Proceedings of the 2021 Conference of the North American
  Chapter of the Association for Computational Linguistics: Human Language
  Technologies}, pages 3167--3185, Online. Association for Computational
  Linguistics.

\bibitem[{Chen et~al.(2021{\natexlab{a}})Chen, Liu, Chen, and
  Zhang}]{chen-etal-2021-dialogsum}
Yulong Chen, Yang Liu, Liang Chen, and Yue Zhang. 2021{\natexlab{a}}.
\newblock \href {https://doi.org/10.18653/v1/2021.findings-acl.449}
  {{D}ialog{S}um: {A} real-life scenario dialogue summarization dataset}.
\newblock In \emph{Findings of the Association for Computational Linguistics:
  ACL-IJCNLP 2021}, pages 5062--5074, Online. Association for Computational
  Linguistics.

\bibitem[{Chen et~al.(2021{\natexlab{b}})Chen, Sadoc, D'Haro, Banchs, and
  Rudnicky}]{chen2021automatic}
Zhang Chen, Jo{\~a}o Sadoc, Luis~Fernando D'Haro, Rafael Banchs, and Alexander
  Rudnicky. 2021{\natexlab{b}}.
\newblock Automatic evaluation and moderation of open-domain dialogue systems.
\newblock \emph{arXiv preprint arXiv:2111.02110}.

\bibitem[{Choi et~al.(2018)Choi, He, Iyyer, Yatskar, Yih, Choi, Liang, and
  Zettlemoyer}]{choi-etal-2018-quac}
Eunsol Choi, He~He, Mohit Iyyer, Mark Yatskar, Wen-tau Yih, Yejin Choi, Percy
  Liang, and Luke Zettlemoyer. 2018.
\newblock \href {https://doi.org/10.18653/v1/D18-1241} {{Q}u{AC}: Question
  answering in context}.
\newblock In \emph{Proceedings of the 2018 Conference on Empirical Methods in
  Natural Language Processing}, pages 2174--2184, Brussels, Belgium.
  Association for Computational Linguistics.

\bibitem[{Coope et~al.(2020{\natexlab{a}})Coope, Farghly, Gerz, Vuli{\'c}, and
  Henderson}]{coope2020span}
Sam Coope, Tyler Farghly, Daniela Gerz, Ivan Vuli{\'c}, and Matthew Henderson.
  2020{\natexlab{a}}.
\newblock Span-convert: Few-shot span extraction for dialog with pretrained
  conversational representations.
\newblock \emph{arXiv preprint arXiv:2005.08866}.

\bibitem[{Coope et~al.(2020{\natexlab{b}})Coope, Farghly, Gerz, Vuli{\'c}, and
  Henderson}]{coope-etal-2020-span}
Samuel Coope, Tyler Farghly, Daniela Gerz, Ivan Vuli{\'c}, and Matthew
  Henderson. 2020{\natexlab{b}}.
\newblock \href {https://doi.org/10.18653/v1/2020.acl-main.11}
  {{S}pan-{ConveRT}: {F}ew-shot span extraction for dialog with pretrained
  conversational representations}.
\newblock In \emph{Proceedings of the 58th Annual Meeting of the Association
  for Computational Linguistics}, pages 107--121, Online. Association for
  Computational Linguistics.

\bibitem[{Coucke et~al.(2018)Coucke, Saade, Ball, Bluche, Caulier, Leroy,
  Doumouro, Gisselbrecht, Caltagirone, Lavril et~al.}]{coucke2018snips}
Alice Coucke, Alaa Saade, Adrien Ball, Th{\'e}odore Bluche, Alexandre Caulier,
  David Leroy, Cl{\'e}ment Doumouro, Thibault Gisselbrecht, Francesco
  Caltagirone, Thibaut Lavril, et~al. 2018.
\newblock Snips voice platform: an embedded spoken language understanding
  system for private-by-design voice interfaces.
\newblock \emph{arXiv preprint arXiv:1805.10190}, pages 12--16.

\bibitem[{Cui et~al.(2020)Cui, Wu, Liu, Zhang, and Zhou}]{cui-etal-2020-mutual}
Leyang Cui, Yu~Wu, Shujie Liu, Yue Zhang, and Ming Zhou. 2020.
\newblock \href {https://doi.org/10.18653/v1/2020.acl-main.130} {{M}u{T}ual: A
  dataset for multi-turn dialogue reasoning}.
\newblock In \emph{Proceedings of the 58th Annual Meeting of the Association
  for Computational Linguistics}, pages 1406--1416, Online. Association for
  Computational Linguistics.

\bibitem[{Demszky et~al.(2020)Demszky, Movshovitz-Attias, Ko, Cowen, Nemade,
  and Ravi}]{demszky-etal-2020-goemotions}
Dorottya Demszky, Dana Movshovitz-Attias, Jeongwoo Ko, Alan Cowen, Gaurav
  Nemade, and Sujith Ravi. 2020.
\newblock \href {https://doi.org/10.18653/v1/2020.acl-main.372}
  {{G}o{E}motions: A dataset of fine-grained emotions}.
\newblock In \emph{Proceedings of the 58th Annual Meeting of the Association
  for Computational Linguistics}, pages 4040--4054, Online. Association for
  Computational Linguistics.

\bibitem[{Devlin et~al.(2019)Devlin, Chang, Lee, and
  Toutanova}]{devlin-etal-2019-bert}
Jacob Devlin, Ming-Wei Chang, Kenton Lee, and Kristina Toutanova. 2019.
\newblock \href {https://doi.org/10.18653/v1/N19-1423} {{BERT}: Pre-training of
  deep bidirectional transformers for language understanding}.
\newblock In \emph{Proceedings of the 2019 Conference of the North {A}merican
  Chapter of the Association for Computational Linguistics: Human Language
  Technologies, Volume 1 (Long and Short Papers)}, pages 4171--4186,
  Minneapolis, Minnesota. Association for Computational Linguistics.

\bibitem[{Dinan et~al.(2019{\natexlab{a}})Dinan, Humeau, Chintagunta, and
  Weston}]{dinan-etal-2019-build}
Emily Dinan, Samuel Humeau, Bharath Chintagunta, and Jason Weston.
  2019{\natexlab{a}}.
\newblock \href {https://doi.org/10.18653/v1/D19-1461} {Build it break it fix
  it for dialogue safety: Robustness from adversarial human attack}.
\newblock In \emph{Proceedings of the 2019 Conference on Empirical Methods in
  Natural Language Processing and the 9th International Joint Conference on
  Natural Language Processing (EMNLP-IJCNLP)}, pages 4537--4546, Hong Kong,
  China. Association for Computational Linguistics.

\bibitem[{Dinan et~al.(2019{\natexlab{b}})Dinan, Logacheva, Malykh, Miller,
  Shuster, Urbanek, Kiela, Szlam, Serban, Lowe et~al.}]{dinan2019second}
Emily Dinan, Varvara Logacheva, Valentin Malykh, Alexander Miller, Kurt
  Shuster, Jack Urbanek, Douwe Kiela, Arthur Szlam, Iulian Serban, Ryan Lowe,
  et~al. 2019{\natexlab{b}}.
\newblock The second conversational intelligence challenge (convai2).
\newblock \emph{arXiv preprint arXiv:1902.00098}.

\bibitem[{Dinan et~al.(2019{\natexlab{c}})Dinan, Roller, Shuster, Fan, Auli,
  and Weston}]{dinan2018wizard}
Emily Dinan, Stephen Roller, Kurt Shuster, Angela Fan, Michael Auli, and Jason
  Weston. 2019{\natexlab{c}}.
\newblock {W}izard of {W}ikipedia: Knowledge-powered conversational agents.
\newblock In \emph{Proceedings of the International Conference on Learning
  Representations (ICLR)}.

\bibitem[{El~Asri et~al.(2017)El~Asri, Schulz, Sharma, Zumer, Harris, Fine,
  Mehrotra, and Suleman}]{el-asri-etal-2017-frames}
Layla El~Asri, Hannes Schulz, Shikhar Sharma, Jeremie Zumer, Justin Harris,
  Emery Fine, Rahul Mehrotra, and Kaheer Suleman. 2017.
\newblock \href {https://doi.org/10.18653/v1/W17-5526} {{F}rames: a corpus for
  adding memory to goal-oriented dialogue systems}.
\newblock In \emph{Proceedings of the 18th Annual {SIG}dial Meeting on
  Discourse and Dialogue}, pages 207--219, Saarbr{\"u}cken, Germany.
  Association for Computational Linguistics.

\bibitem[{Eric et~al.(2017)Eric, Krishnan, Charette, and
  Manning}]{eric-etal-2017-key}
Mihail Eric, Lakshmi Krishnan, Francois Charette, and Christopher~D. Manning.
  2017.
\newblock \href {https://doi.org/10.18653/v1/W17-5506} {Key-value retrieval
  networks for task-oriented dialogue}.
\newblock In \emph{Proceedings of the 18th Annual {SIG}dial Meeting on
  Discourse and Dialogue}, pages 37--49, Saarbr{\"u}cken, Germany. Association
  for Computational Linguistics.

\bibitem[{Feng et~al.(2020{\natexlab{a}})Feng, Wan, Gunasekara, Patel, Joshi,
  and Lastras}]{feng-etal-2020-doc2dial}
Song Feng, Hui Wan, Chulaka Gunasekara, Siva Patel, Sachindra Joshi, and Luis
  Lastras. 2020{\natexlab{a}}.
\newblock \href {https://doi.org/10.18653/v1/2020.emnlp-main.652} {doc2dial: A
  goal-oriented document-grounded dialogue dataset}.
\newblock In \emph{Proceedings of the 2020 Conference on Empirical Methods in
  Natural Language Processing (EMNLP)}, pages 8118--8128, Online. Association
  for Computational Linguistics.

\bibitem[{Feng et~al.(2020{\natexlab{b}})Feng, Mehri, Eskenazi, and
  Zhao}]{feng-etal-2020-none}
Yulan Feng, Shikib Mehri, Maxine Eskenazi, and Tiancheng Zhao.
  2020{\natexlab{b}}.
\newblock \href {https://doi.org/10.18653/v1/2020.acl-main.182} {{``}none of
  the above{''}: Measure uncertainty in dialog response retrieval}.
\newblock In \emph{Proceedings of the 58th Annual Meeting of the Association
  for Computational Linguistics}, pages 2013--2020, Online. Association for
  Computational Linguistics.

\bibitem[{Galley et~al.(2019)Galley, Brockett, Gao, Gao, and
  Dolan}]{Galley2019GroundedRG}
Michel Galley, Chris Brockett, Xiang Gao, Jianfeng Gao, and Bill Dolan. 2019.
\newblock Grounded response generation task at dstc7.

\bibitem[{Gao et~al.(2020)Gao, Zhang, Galley, Brockett, and
  Dolan}]{gao-etal-2020-dialogue}
Xiang Gao, Yizhe Zhang, Michel Galley, Chris Brockett, and Bill Dolan. 2020.
\newblock \href {https://doi.org/10.18653/v1/2020.emnlp-main.28} {Dialogue
  response ranking training with large-scale human feedback data}.
\newblock In \emph{Proceedings of the 2020 Conference on Empirical Methods in
  Natural Language Processing (EMNLP)}, pages 386--395, Online. Association for
  Computational Linguistics.

\bibitem[{Gliwa et~al.(2019)Gliwa, Mochol, Biesek, and Wawer}]{gliwa2019samsum}
Bogdan Gliwa, Iwona Mochol, Maciej Biesek, and Aleksander Wawer. 2019.
\newblock Samsum corpus: A human-annotated dialogue dataset for abstractive
  summarization.
\newblock \emph{arXiv preprint arXiv:1911.12237}.

\bibitem[{Goo and Chen(2018)}]{goo2018abstractive}
Chih-Wen Goo and Yun-Nung Chen. 2018.
\newblock Abstractive dialogue summarization with sentence-gated modeling
  optimized by dialogue acts.
\newblock In \emph{Proceedings of 7th IEEE Workshop on Spoken Language
  Technology}.

\bibitem[{Gopalakrishnan et~al.(2019)Gopalakrishnan, Hedayatnia, Chen,
  Gottardi, Kwatra, Venkatesh, Gabriel, and Hakkani-Tür}]{Gopalakrishnan2019}
Karthik Gopalakrishnan, Behnam Hedayatnia, Qinlang Chen, Anna Gottardi, Sanjeev
  Kwatra, Anu Venkatesh, Raefer Gabriel, and Dilek Hakkani-Tür. 2019.
\newblock \href {https://doi.org/10.21437/Interspeech.2019-3079}
  {{Topical-Chat: Towards Knowledge-Grounded Open-Domain Conversations}}.
\newblock In \emph{Proc. Interspeech 2019}, pages 1891--1895.

\bibitem[{Gupta et~al.(2019)Gupta, Mehri, Zhao, Pavel, Eskenazi, and
  Bigham}]{gupta-etal-2019-investigating}
Prakhar Gupta, Shikib Mehri, Tiancheng Zhao, Amy Pavel, Maxine Eskenazi, and
  Jeffrey Bigham. 2019.
\newblock \href {https://doi.org/10.18653/v1/W19-5944} {Investigating
  evaluation of open-domain dialogue systems with human generated multiple
  references}.
\newblock In \emph{Proceedings of the 20th Annual SIGdial Meeting on Discourse
  and Dialogue}, pages 379--391, Stockholm, Sweden. Association for
  Computational Linguistics.

\bibitem[{Gupta et~al.(2021)Gupta, Wu, Liu, and Xiong}]{gupta2021dialfact}
Prakhar Gupta, Chien-Sheng Wu, Wenhao Liu, and Caiming Xiong. 2021.
\newblock Dialfact: A benchmark for fact-checking in dialogue.
\newblock \emph{arXiv preprint arXiv:2110.08222}.

\bibitem[{Ham et~al.(2020)Ham, Lee, Jang, and Kim}]{ham-etal-2020-end}
Donghoon Ham, Jeong-Gwan Lee, Youngsoo Jang, and Kee-Eung Kim. 2020.
\newblock \href {https://doi.org/10.18653/v1/2020.acl-main.54} {End-to-end
  neural pipeline for goal-oriented dialogue systems using {GPT}-2}.
\newblock In \emph{Proceedings of the 58th Annual Meeting of the Association
  for Computational Linguistics}, pages 583--592, Online. Association for
  Computational Linguistics.

\bibitem[{Hemphill et~al.(1990)Hemphill, Godfrey, and
  Doddington}]{hemphill-etal-1990-atis}
Charles~T. Hemphill, John~J. Godfrey, and George~R. Doddington. 1990.
\newblock \href {https://aclanthology.org/H90-1021} {The {ATIS} spoken language
  systems pilot corpus}.
\newblock In \emph{Speech and Natural Language: Proceedings of a Workshop Held
  at Hidden Valley, {P}ennsylvania, June 24-27,1990}.

\bibitem[{Henderson and Vuli{\'c}(2020)}]{henderson2020convex}
Matthew Henderson and Ivan Vuli{\'c}. 2020.
\newblock Convex: Data-efficient and few-shot slot labeling.
\newblock \emph{arXiv preprint arXiv:2010.11791}.

\bibitem[{Hori and Hori(2017)}]{DSTC6_End-to-End_Conversation_Modeling}
Chiori Hori and Takaaki Hori. 2017.
\newblock End-to-end conversation modeling track in dstc6.
\newblock \emph{arXiv:1706.07440}.

\bibitem[{Hosseini-Asl et~al.(2020)Hosseini-Asl, McCann, Wu, Yavuz, and
  Socher}]{hosseini2020simple}
Ehsan Hosseini-Asl, Bryan McCann, Chien-Sheng Wu, Semih Yavuz, and Richard
  Socher. 2020.
\newblock A simple language model for task-oriented dialogue.
\newblock \emph{Advances in Neural Information Processing Systems},
  33:20179--20191.

\bibitem[{Hsu et~al.(2018)Hsu, Chen, Kuo, Huang, and
  Ku}]{hsu-etal-2018-emotionlines}
Chao-Chun Hsu, Sheng-Yeh Chen, Chuan-Chun Kuo, Ting-Hao Huang, and Lun-Wei Ku.
  2018.
\newblock \href {https://aclanthology.org/L18-1252} {{E}motion{L}ines: An
  emotion corpus of multi-party conversations}.
\newblock In \emph{Proceedings of the Eleventh International Conference on
  Language Resources and Evaluation ({LREC} 2018)}, Miyazaki, Japan. European
  Language Resources Association (ELRA).

\bibitem[{Huang et~al.(2020)Huang, Ye, Qin, Lin, and
  Liang}]{huang-etal-2020-grade}
Lishan Huang, Zheng Ye, Jinghui Qin, Liang Lin, and Xiaodan Liang. 2020.
\newblock \href {https://doi.org/10.18653/v1/2020.emnlp-main.742} {{GRADE}:
  Automatic graph-enhanced coherence metric for evaluating open-domain dialogue
  systems}.
\newblock In \emph{Proceedings of the 2020 Conference on Empirical Methods in
  Natural Language Processing (EMNLP)}, pages 9230--9240, Online. Association
  for Computational Linguistics.

\bibitem[{Kingma and Ba(2015)}]{kingma:adam}
Diederick~P Kingma and Jimmy Ba. 2015.
\newblock Adam: A method for stochastic optimization.
\newblock In \emph{International Conference on Learning Representations
  (ICLR)}.

\bibitem[{Larson and Leach(2022)}]{larson2022redwood}
Stefan Larson and Kevin Leach. 2022.
\newblock Redwood: Using collision detection to grow a large-scale intent
  classification dataset.
\newblock \emph{arXiv preprint arXiv:2204.05483}.

\bibitem[{Larson et~al.(2019)Larson, Mahendran, Peper, Clarke, Lee, Hill,
  Kummerfeld, Leach, Laurenzano, Tang, and Mars}]{larson-etal-2019-evaluation}
Stefan Larson, Anish Mahendran, Joseph~J. Peper, Christopher Clarke, Andrew
  Lee, Parker Hill, Jonathan~K. Kummerfeld, Kevin Leach, Michael~A. Laurenzano,
  Lingjia Tang, and Jason Mars. 2019.
\newblock \href {https://doi.org/10.18653/v1/D19-1131} {An evaluation dataset
  for intent classification and out-of-scope prediction}.
\newblock In \emph{Proceedings of the 2019 Conference on Empirical Methods in
  Natural Language Processing and the 9th International Joint Conference on
  Natural Language Processing (EMNLP-IJCNLP)}, pages 1311--1316, Hong Kong,
  China. Association for Computational Linguistics.

\bibitem[{Lester et~al.(2021)Lester, Al-Rfou, and
  Constant}]{lester-etal-2021-power}
Brian Lester, Rami Al-Rfou, and Noah Constant. 2021.
\newblock \href {https://doi.org/10.18653/v1/2021.emnlp-main.243} {The power of
  scale for parameter-efficient prompt tuning}.
\newblock In \emph{Proceedings of the 2021 Conference on Empirical Methods in
  Natural Language Processing}, pages 3045--3059, Online and Punta Cana,
  Dominican Republic. Association for Computational Linguistics.

\bibitem[{Lewis et~al.(2020)Lewis, Liu, Goyal, Ghazvininejad, Mohamed, Levy,
  Stoyanov, and Zettlemoyer}]{lewis-etal-2020-bart}
Mike Lewis, Yinhan Liu, Naman Goyal, Marjan Ghazvininejad, Abdelrahman Mohamed,
  Omer Levy, Veselin Stoyanov, and Luke Zettlemoyer. 2020.
\newblock \href {https://doi.org/10.18653/v1/2020.acl-main.703} {{BART}:
  Denoising sequence-to-sequence pre-training for natural language generation,
  translation, and comprehension}.
\newblock In \emph{Proceedings of the 58th Annual Meeting of the Association
  for Computational Linguistics}, pages 7871--7880, Online. Association for
  Computational Linguistics.

\bibitem[{Lewis et~al.(2017)Lewis, Yarats, Dauphin, Parikh, and
  Batra}]{lewis-etal-2017-deal}
Mike Lewis, Denis Yarats, Yann Dauphin, Devi Parikh, and Dhruv Batra. 2017.
\newblock \href {https://doi.org/10.18653/v1/D17-1259} {Deal or no deal?
  end-to-end learning of negotiation dialogues}.
\newblock In \emph{Proceedings of the 2017 Conference on Empirical Methods in
  Natural Language Processing}, pages 2443--2453, Copenhagen, Denmark.
  Association for Computational Linguistics.

\bibitem[{Li et~al.(2018)Li, Panda, Liu, and Gao}]{li2018microsoft}
Xiujun Li, Sarah Panda, Jingjing Liu, and Jianfeng Gao. 2018.
\newblock Microsoft dialogue challenge: Building end-to-end task-completion
  dialogue systems.
\newblock \emph{arXiv preprint arXiv:1807.11125}.

\bibitem[{Li et~al.(2017)Li, Su, Shen, Li, Cao, and
  Niu}]{li-etal-2017-dailydialog}
Yanran Li, Hui Su, Xiaoyu Shen, Wenjie Li, Ziqiang Cao, and Shuzi Niu. 2017.
\newblock \href {https://aclanthology.org/I17-1099} {{D}aily{D}ialog: A
  manually labelled multi-turn dialogue dataset}.
\newblock In \emph{Proceedings of the Eighth International Joint Conference on
  Natural Language Processing (Volume 1: Long Papers)}, pages 986--995, Taipei,
  Taiwan. Asian Federation of Natural Language Processing.

\bibitem[{Li et~al.(2021)Li, Zhang, Fei, Feng, and
  Zhou}]{li-etal-2021-conversations}
Zekang Li, Jinchao Zhang, Zhengcong Fei, Yang Feng, and Jie Zhou. 2021.
\newblock \href {https://doi.org/10.18653/v1/2021.acl-long.11} {Conversations
  are not flat: Modeling the dynamic information flow across dialogue
  utterances}.
\newblock In \emph{Proceedings of the 59th Annual Meeting of the Association
  for Computational Linguistics and the 11th International Joint Conference on
  Natural Language Processing (Volume 1: Long Papers)}, pages 128--138, Online.
  Association for Computational Linguistics.

\bibitem[{Lin et~al.(2022)Lin, Tan, Miller, Tian, and
  Ren}]{Lin2022UnsupervisedCG}
Bill~Yuchen Lin, Kangmin Tan, Chris Miller, Beiwen Tian, and Xiang Ren. 2022.
\newblock Unsupervised cross-task generalization via retrieval augmentation.
\newblock \emph{ArXiv}, abs/2204.07937.

\bibitem[{Lin et~al.(2020)Lin, Madotto, Winata, and Fung}]{lin-etal-2020-mintl}
Zhaojiang Lin, Andrea Madotto, Genta~Indra Winata, and Pascale Fung. 2020.
\newblock \href {https://doi.org/10.18653/v1/2020.emnlp-main.273} {{M}in{TL}:
  Minimalist transfer learning for task-oriented dialogue systems}.
\newblock In \emph{Proceedings of the 2020 Conference on Empirical Methods in
  Natural Language Processing (EMNLP)}, pages 3391--3405, Online. Association
  for Computational Linguistics.

\bibitem[{Liu et~al.(2021{\natexlab{a}})Liu, Yuan, Fu, Jiang, Hayashi, and
  Neubig}]{liu2021pre}
Pengfei Liu, Weizhe Yuan, Jinlan Fu, Zhengbao Jiang, Hiroaki Hayashi, and
  Graham Neubig. 2021{\natexlab{a}}.
\newblock Pre-train, prompt, and predict: A systematic survey of prompting
  methods in natural language processing.
\newblock \emph{arXiv preprint arXiv:2107.13586}.

\bibitem[{Liu et~al.(2022)Liu, Yu, Rimell, and Blunsom}]{TACL2889}
Qi~Liu, Lei Yu, Laura Rimell, and Phil Blunsom. 2022.
\newblock \href {https://transacl.org/ojs/index.php/tacl/article/view/2889}
  {Pretraining the noisy channel model for task-oriented dialogue}.
\newblock \emph{Transactions of the Association for Computational Linguistics},
  9(0):657--674.

\bibitem[{Liu et~al.(2021{\natexlab{b}})Liu, Eshghi, Swietojanski, and
  Rieser}]{liu2021benchmarking}
Xingkun Liu, Arash Eshghi, Pawel Swietojanski, and Verena Rieser.
  2021{\natexlab{b}}.
\newblock Benchmarking natural language understanding services for building
  conversational agents.
\newblock In \emph{Increasing Naturalness and Flexibility in Spoken Dialogue
  Interaction}, pages 165--183. Springer.

\bibitem[{Liu et~al.(2020)Liu, Winata, Xu, and Fung}]{liu2020coach}
Zihan Liu, Genta~Indra Winata, Peng Xu, and Pascale Fung. 2020.
\newblock Coach: A coarse-to-fine approach for cross-domain slot filling.
\newblock \emph{arXiv preprint arXiv:2004.11727}.

\bibitem[{Madotto et~al.(2021)Madotto, Lin, Winata, and Fung}]{madotto2021few}
Andrea Madotto, Zhaojiang Lin, Genta~Indra Winata, and Pascale Fung. 2021.
\newblock Few-shot bot: Prompt-based learning for dialogue systems.
\newblock \emph{arXiv preprint arXiv:2110.08118}.

\bibitem[{Mehri and Eric(2021)}]{mehri-eric-2021-example}
Shikib Mehri and Mihail Eric. 2021.
\newblock \href {https://doi.org/10.18653/v1/2021.naacl-main.237}
  {Example-driven intent prediction with observers}.
\newblock In \emph{Proceedings of the 2021 Conference of the North American
  Chapter of the Association for Computational Linguistics: Human Language
  Technologies}, pages 2979--2992, Online. Association for Computational
  Linguistics.

\bibitem[{Mehri and
  Eskenazi(2020{\natexlab{a}})}]{mehri-eskenazi-2020-unsupervised}
Shikib Mehri and Maxine Eskenazi. 2020{\natexlab{a}}.
\newblock \href {https://aclanthology.org/2020.sigdial-1.28} {Unsupervised
  evaluation of interactive dialog with {D}ialo{GPT}}.
\newblock In \emph{Proceedings of the 21th Annual Meeting of the Special
  Interest Group on Discourse and Dialogue}, pages 225--235, 1st virtual
  meeting. Association for Computational Linguistics.

\bibitem[{Mehri and Eskenazi(2020{\natexlab{b}})}]{mehri-eskenazi-2020-usr}
Shikib Mehri and Maxine Eskenazi. 2020{\natexlab{b}}.
\newblock \href {https://doi.org/10.18653/v1/2020.acl-main.64} {{USR}: An
  unsupervised and reference free evaluation metric for dialog generation}.
\newblock In \emph{Proceedings of the 58th Annual Meeting of the Association
  for Computational Linguistics}, pages 681--707, Online. Association for
  Computational Linguistics.

\bibitem[{Mehri and Eskenazi(2021)}]{mehri2021gensf}
Shikib Mehri and Maxine Eskenazi. 2021.
\newblock Gensf: Simultaneous adaptation of generative pre-trained models and
  slot filling.
\newblock \emph{arXiv preprint arXiv:2106.07055}.

\bibitem[{Mehri et~al.(2019)Mehri, Razumovskaia, Zhao, and
  Eskenazi}]{mehri-etal-2019-pretraining}
Shikib Mehri, Evgeniia Razumovskaia, Tiancheng Zhao, and Maxine Eskenazi. 2019.
\newblock \href {https://doi.org/10.18653/v1/P19-1373} {Pretraining methods for
  dialog context representation learning}.
\newblock In \emph{Proceedings of the 57th Annual Meeting of the Association
  for Computational Linguistics}, pages 3836--3845, Florence, Italy.
  Association for Computational Linguistics.

\bibitem[{Merdivan et~al.(2020)Merdivan, Singh, Hanke, Kropf, Holzinger, and
  Geist}]{merdivan2020human}
Erinc Merdivan, Deepika Singh, Sten Hanke, Johannes Kropf, Andreas Holzinger,
  and Matthieu Geist. 2020.
\newblock Human annotated dialogues dataset for natural conversational agents.
\newblock \emph{Applied Sciences}, 10(3):762.

\bibitem[{Mi et~al.(2021{\natexlab{a}})Mi, Li, Wang, Jiang, and
  Liu}]{Mi2021CINSCI}
Fei Mi, Yitong Li, Yasheng Wang, Xin Jiang, and Qun Liu. 2021{\natexlab{a}}.
\newblock Cins: Comprehensive instruction for few-shot learning in
  task-oriented dialog systems.
\newblock \emph{ArXiv}, abs/2109.04645.

\bibitem[{Mi et~al.(2021{\natexlab{b}})Mi, Zhou, Kong, Cai, Huang, and
  Faltings}]{mi-etal-2021-self}
Fei Mi, Wanhao Zhou, Lingjing Kong, Fengyu Cai, Minlie Huang, and Boi Faltings.
  2021{\natexlab{b}}.
\newblock \href {https://doi.org/10.18653/v1/2021.emnlp-main.142}
  {Self-training improves pre-training for few-shot learning in task-oriented
  dialog systems}.
\newblock In \emph{Proceedings of the 2021 Conference on Empirical Methods in
  Natural Language Processing}, pages 1887--1898, Online and Punta Cana,
  Dominican Republic. Association for Computational Linguistics.

\bibitem[{Min et~al.(2022{\natexlab{a}})Min, Lewis, Zettlemoyer, and
  Hajishirzi}]{min2022metaicl}
Sewon Min, Mike Lewis, Luke Zettlemoyer, and Hannaneh Hajishirzi.
  2022{\natexlab{a}}.
\newblock Meta{ICL}: Learning to learn in context.
\newblock In \emph{NAACL-HLT}.

\bibitem[{Min et~al.(2022{\natexlab{b}})Min, Lyu, Holtzman, Artetxe, Lewis,
  Hajishirzi, and Zettlemoyer}]{min2022rethinking}
Sewon Min, Xinxi Lyu, Ari Holtzman, Mikel Artetxe, Mike Lewis, Hannaneh
  Hajishirzi, and Luke Zettlemoyer. 2022{\natexlab{b}}.
\newblock Rethinking the role of demonstrations: What makes in-context learning
  work?
\newblock \emph{arXiv preprint arXiv:2202.12837}.

\bibitem[{Mishra et~al.(2021)Mishra, Khashabi, Baral, and
  Hajishirzi}]{mishra2021natural}
Swaroop Mishra, Daniel Khashabi, Chitta Baral, and Hannaneh Hajishirzi. 2021.
\newblock Natural instructions: Benchmarking generalization to new tasks from
  natural language instructions.
\newblock \emph{In Annual Meeting of the Association for Computational
  Linguistics (ACL)}.

\bibitem[{Moon et~al.(2019)Moon, Shah, Kumar, and
  Subba}]{moon-etal-2019-opendialkg}
Seungwhan Moon, Pararth Shah, Anuj Kumar, and Rajen Subba. 2019.
\newblock \href {https://doi.org/10.18653/v1/P19-1081} {{O}pen{D}ial{KG}:
  Explainable conversational reasoning with attention-based walks over
  knowledge graphs}.
\newblock In \emph{Proceedings of the 57th Annual Meeting of the Association
  for Computational Linguistics}, pages 845--854, Florence, Italy. Association
  for Computational Linguistics.

\bibitem[{Mrk{\v{s}}i{\'c} et~al.(2017)Mrk{\v{s}}i{\'c}, {\'O}~S{\'e}aghdha,
  Wen, Thomson, and Young}]{mrksic-etal-2017-neural}
Nikola Mrk{\v{s}}i{\'c}, Diarmuid {\'O}~S{\'e}aghdha, Tsung-Hsien Wen, Blaise
  Thomson, and Steve Young. 2017.
\newblock \href {https://doi.org/10.18653/v1/P17-1163} {Neural belief tracker:
  Data-driven dialogue state tracking}.
\newblock In \emph{Proceedings of the 55th Annual Meeting of the Association
  for Computational Linguistics (Volume 1: Long Papers)}, pages 1777--1788,
  Vancouver, Canada. Association for Computational Linguistics.

\bibitem[{Nie et~al.(2021)Nie, Williamson, Bansal, Kiela, and
  Weston}]{nie-etal-2021-like}
Yixin Nie, Mary Williamson, Mohit Bansal, Douwe Kiela, and Jason Weston. 2021.
\newblock \href {https://doi.org/10.18653/v1/2021.acl-long.134} {{I} like fish,
  especially dolphins: Addressing contradictions in dialogue modeling}.
\newblock In \emph{Proceedings of the 59th Annual Meeting of the Association
  for Computational Linguistics and the 11th International Joint Conference on
  Natural Language Processing (Volume 1: Long Papers)}, pages 1699--1713,
  Online. Association for Computational Linguistics.

\bibitem[{Ouyang et~al.(2022)Ouyang, Wu, Jiang, Almeida, Wainwright, Mishkin,
  Zhang, Agarwal, Slama, Ray et~al.}]{ouyang2022training}
Long Ouyang, Jeff Wu, Xu~Jiang, Diogo Almeida, Carroll~L Wainwright, Pamela
  Mishkin, Chong Zhang, Sandhini Agarwal, Katarina Slama, Alex Ray, et~al.
  2022.
\newblock Training language models to follow instructions with human feedback.
\newblock \emph{arXiv preprint arXiv:2203.02155}.

\bibitem[{Papineni et~al.(2002)Papineni, Roukos, Ward, and
  Zhu}]{papineni2002bleu}
Kishore Papineni, Salim Roukos, Todd Ward, and Wei-Jing Zhu. 2002.
\newblock Bleu: a method for automatic evaluation of machine translation.
\newblock In \emph{Proceedings of the 40th annual meeting of the Association
  for Computational Linguistics}, pages 311--318.

\bibitem[{Peng et~al.(2021)Peng, Li, Li, Shayandeh, Liden, and
  Gao}]{peng2021soloist}
Baolin Peng, Chunyuan Li, Jinchao Li, Shahin Shayandeh, Lars Liden, and
  Jianfeng Gao. 2021.
\newblock \href
  {https://www.microsoft.com/en-us/research/publication/soloist-building-task-bots-at-scale-with-transfer-learning-and-machine-teaching/}
  {Soloist: Building task bots at scale with transfer learning and machine
  teaching}.
\newblock In \emph{Transactions of the Association for Computational
  Linguistics}.

\bibitem[{Phy et~al.(2020)Phy, Zhao, and Aizawa}]{phy-etal-2020-deconstruct}
Vitou Phy, Yang Zhao, and Akiko Aizawa. 2020.
\newblock \href {https://doi.org/10.18653/v1/2020.coling-main.368} {Deconstruct
  to reconstruct a configurable evaluation metric for open-domain dialogue
  systems}.
\newblock In \emph{Proceedings of the 28th International Conference on
  Computational Linguistics}, pages 4164--4178, Barcelona, Spain (Online).
  International Committee on Computational Linguistics.

\bibitem[{Qin et~al.(2021)Qin, Gupta, Upadhyay, He, Choi, and
  Faruqui}]{qin-etal-2021-timedial}
Lianhui Qin, Aditya Gupta, Shyam Upadhyay, Luheng He, Yejin Choi, and Manaal
  Faruqui. 2021.
\newblock \href {https://doi.org/10.18653/v1/2021.acl-long.549} {{TIMEDIAL}:
  Temporal commonsense reasoning in dialog}.
\newblock In \emph{Proceedings of the 59th Annual Meeting of the Association
  for Computational Linguistics and the 11th International Joint Conference on
  Natural Language Processing (Volume 1: Long Papers)}, pages 7066--7076,
  Online. Association for Computational Linguistics.

\bibitem[{Radford et~al.(2019)Radford, Wu, Child, Luan, Amodei, and
  Sutskever}]{Radford2019LanguageMA}
Alec Radford, Jeff Wu, Rewon Child, David Luan, Dario Amodei, and Ilya
  Sutskever. 2019.
\newblock Language models are unsupervised multitask learners.

\bibitem[{Raghu et~al.(2021)Raghu, Agarwal, Joshi, and
  {Mausam}}]{raghu-etal-2021-end}
Dinesh Raghu, Shantanu Agarwal, Sachindra Joshi, and {Mausam}. 2021.
\newblock \href {https://doi.org/10.18653/v1/2021.emnlp-main.357} {End-to-end
  learning of flowchart grounded task-oriented dialogs}.
\newblock In \emph{Proceedings of the 2021 Conference on Empirical Methods in
  Natural Language Processing}, pages 4348--4366, Online and Punta Cana,
  Dominican Republic. Association for Computational Linguistics.

\bibitem[{Rashkin et~al.(2019)Rashkin, Smith, Li, and
  Boureau}]{rashkin-etal-2019-towards}
Hannah Rashkin, Eric~Michael Smith, Margaret Li, and Y-Lan Boureau. 2019.
\newblock \href {https://doi.org/10.18653/v1/P19-1534} {Towards empathetic
  open-domain conversation models: A new benchmark and dataset}.
\newblock In \emph{Proceedings of the 57th Annual Meeting of the Association
  for Computational Linguistics}, pages 5370--5381, Florence, Italy.
  Association for Computational Linguistics.

\bibitem[{Rastogi et~al.(2020{\natexlab{a}})Rastogi, Zang, Sunkara, Gupta, and
  Khaitan}]{rastogi2020schema}
Abhinav Rastogi, Xiaoxue Zang, Srinivas Sunkara, Raghav Gupta, and Pranav
  Khaitan. 2020{\natexlab{a}}.
\newblock Schema-guided dialogue state tracking task at dstc8.
\newblock \emph{arXiv preprint arXiv:2002.01359}.

\bibitem[{Rastogi et~al.(2020{\natexlab{b}})Rastogi, Zang, Sunkara, Gupta, and
  Khaitan}]{rastogi2020towards}
Abhinav Rastogi, Xiaoxue Zang, Srinivas Sunkara, Raghav Gupta, and Pranav
  Khaitan. 2020{\natexlab{b}}.
\newblock Towards scalable multi-domain conversational agents: The
  schema-guided dialogue dataset.
\newblock In \emph{Proceedings of the AAAI Conference on Artificial
  Intelligence}, volume~34, pages 8689--8696.

\bibitem[{Reddy et~al.(2019)Reddy, Chen, and Manning}]{reddy-etal-2019-coqa}
Siva Reddy, Danqi Chen, and Christopher~D. Manning. 2019.
\newblock \href {https://doi.org/10.1162/tacl_a_00266} {{C}o{QA}: A
  conversational question answering challenge}.
\newblock \emph{Transactions of the Association for Computational Linguistics},
  7:249--266.

\bibitem[{Roller et~al.(2021)Roller, Dinan, Goyal, Ju, Williamson, Liu, Xu,
  Ott, Smith, Boureau, and Weston}]{roller-etal-2021-recipes}
Stephen Roller, Emily Dinan, Naman Goyal, Da~Ju, Mary Williamson, Yinhan Liu,
  Jing Xu, Myle Ott, Eric~Michael Smith, Y-Lan Boureau, and Jason Weston. 2021.
\newblock \href {https://doi.org/10.18653/v1/2021.eacl-main.24} {Recipes for
  building an open-domain chatbot}.
\newblock In \emph{Proceedings of the 16th Conference of the European Chapter
  of the Association for Computational Linguistics: Main Volume}, pages
  300--325, Online. Association for Computational Linguistics.

\bibitem[{Sai et~al.(2020)Sai, Mohankumar, Arora, and
  Khapra}]{sai-etal-2020-improving}
Ananya~B. Sai, Akash~Kumar Mohankumar, Siddhartha Arora, and Mitesh~M. Khapra.
  2020.
\newblock \href {https://doi.org/10.1162/tacl_a_00347} {Improving dialog
  evaluation with a multi-reference adversarial dataset and large scale
  pretraining}.
\newblock \emph{Transactions of the Association for Computational Linguistics},
  8:810--827.

\bibitem[{Sanh et~al.(2022)Sanh, Webson, Raffel, Bach, Sutawika, Alyafeai,
  Chaffin, Stiegler, Scao, Raja, Dey, Bari, Xu, Thakker, Sharma, Szczechla,
  Kim, Chhablani, Nayak, Datta, Chang, Jiang, Wang, Manica, Shen, Yong, Pandey,
  Bawden, Wang, Neeraj, Rozen, Sharma, Santilli, Fevry, Fries, Teehan,
  Biderman, Gao, Bers, Wolf, and Rush}]{sanh:iclr22}
Victor Sanh, Albert Webson, Colin Raffel, Stephen~H. Bach, Lintang Sutawika,
  Zaid Alyafeai, Antoine Chaffin, Arnaud Stiegler, Teven~Le Scao, Arun Raja,
  Manan Dey, M~Saiful Bari, Canwen Xu, Urmish Thakker, Shanya~Sharma Sharma,
  Eliza Szczechla, Taewoon Kim, Gunjan Chhablani, Nihal Nayak, Debajyoti Datta,
  Jonathan Chang, Mike Tian-Jian Jiang, Han Wang, Matteo Manica, Sheng Shen,
  Zheng~Xin Yong, Harshit Pandey, Rachel Bawden, Thomas Wang, Trishala Neeraj,
  Jos Rozen, Abheesht Sharma, Andrea Santilli, Thibault Fevry, Jason~Alan
  Fries, Ryan Teehan, Stella Biderman, Leo Gao, Tali Bers, Thomas Wolf, and
  Alexander~M. Rush. 2022.
\newblock Multitask prompted training enables zero-shot task generalization.
\newblock In \emph{International Conference on Learning Representations
  (ICLR)}.

\bibitem[{Schick and Sch{\"u}tze(2021)}]{schick-schutze-2021-shot}
Timo Schick and Hinrich Sch{\"u}tze. 2021.
\newblock \href {https://doi.org/10.18653/v1/2021.emnlp-main.32} {Few-shot text
  generation with natural language instructions}.
\newblock In \emph{Proceedings of the 2021 Conference on Empirical Methods in
  Natural Language Processing}, pages 390--402, Online and Punta Cana,
  Dominican Republic. Association for Computational Linguistics.

\bibitem[{Scialom et~al.(2021)Scialom, Dray, Lamprier, Piwowarski, Staiano,
  Wang, and Gallinari}]{scialom-etal-2021-questeval}
Thomas Scialom, Paul-Alexis Dray, Sylvain Lamprier, Benjamin Piwowarski, Jacopo
  Staiano, Alex Wang, and Patrick Gallinari. 2021.
\newblock \href {https://doi.org/10.18653/v1/2021.emnlp-main.529}
  {{Q}uest{E}val: Summarization asks for fact-based evaluation}.
\newblock In \emph{Proceedings of the 2021 Conference on Empirical Methods in
  Natural Language Processing}, pages 6594--6604, Online and Punta Cana,
  Dominican Republic. Association for Computational Linguistics.

\bibitem[{See et~al.(2019)See, Roller, Kiela, and Weston}]{see-etal-2019-makes}
Abigail See, Stephen Roller, Douwe Kiela, and Jason Weston. 2019.
\newblock \href {https://doi.org/10.18653/v1/N19-1170} {What makes a good
  conversation? how controllable attributes affect human judgments}.
\newblock In \emph{Proceedings of the 2019 Conference of the North {A}merican
  Chapter of the Association for Computational Linguistics: Human Language
  Technologies, Volume 1 (Long and Short Papers)}, pages 1702--1723,
  Minneapolis, Minnesota. Association for Computational Linguistics.

\bibitem[{Sevegnani et~al.(2021)Sevegnani, Howcroft, Konstas, and
  Rieser}]{sevegnani-etal-2021-otters}
Karin Sevegnani, David~M. Howcroft, Ioannis Konstas, and Verena Rieser. 2021.
\newblock \href {https://doi.org/10.18653/v1/2021.acl-long.194} {{OTT}ers:
  One-turn topic transitions for open-domain dialogue}.
\newblock In \emph{Proceedings of the 59th Annual Meeting of the Association
  for Computational Linguistics and the 11th International Joint Conference on
  Natural Language Processing (Volume 1: Long Papers)}, pages 2492--2504,
  Online. Association for Computational Linguistics.

\bibitem[{Sinha et~al.(2020)Sinha, Parthasarathi, Wang, Lowe, Hamilton, and
  Pineau}]{sinha-etal-2020-learning}
Koustuv Sinha, Prasanna Parthasarathi, Jasmine Wang, Ryan Lowe, William~L.
  Hamilton, and Joelle Pineau. 2020.
\newblock \href {https://doi.org/10.18653/v1/2020.acl-main.220} {Learning an
  unreferenced metric for online dialogue evaluation}.
\newblock In \emph{Proceedings of the 58th Annual Meeting of the Association
  for Computational Linguistics}, pages 2430--2441, Online. Association for
  Computational Linguistics.

\bibitem[{Su et~al.(2022{\natexlab{a}})Su, Shu, Mansimov, Gupta, Cai, Lai, and
  Zhang}]{su2021multitask}
Yixuan Su, Lei Shu, Elman Mansimov, Arshit Gupta, Deng Cai, Yi{-}An Lai, and
  Yi~Zhang. 2022{\natexlab{a}}.
\newblock \href {https://arxiv.org/abs/2109.14739} {Multi-task pre-training for
  plug-and-play task-oriented dialogue system}.

\bibitem[{Su et~al.(2022{\natexlab{b}})Su, Shu, Mansimov, Gupta, Cai, Lai, and
  Zhang}]{su-etal-2022-multi}
Yixuan Su, Lei Shu, Elman Mansimov, Arshit Gupta, Deng Cai, Yi-An Lai, and
  Yi~Zhang. 2022{\natexlab{b}}.
\newblock \href {https://aclanthology.org/2022.acl-long.319} {Multi-task
  pre-training for plug-and-play task-oriented dialogue system}.
\newblock In \emph{Proceedings of the 60th Annual Meeting of the Association
  for Computational Linguistics (Volume 1: Long Papers)}, pages 4661--4676,
  Dublin, Ireland. Association for Computational Linguistics.

\bibitem[{Ung et~al.(2021)Ung, Xu, and Boureau}]{ung2021saferdialogues}
Megan Ung, Jing Xu, and Y-Lan Boureau. 2021.
\newblock \href {http://arxiv.org/abs/2110.07518} {Saferdialogues: Taking
  feedback gracefully after conversational safety failures}.

\bibitem[{Vedantam et~al.(2015)Vedantam, Lawrence~Zitnick, and
  Parikh}]{Vedantam_2015_CVPR}
Ramakrishna Vedantam, C.~Lawrence~Zitnick, and Devi Parikh. 2015.
\newblock Cider: Consensus-based image description evaluation.
\newblock In \emph{The IEEE Conference on Computer Vision and Pattern
  Recognition (CVPR)}.

\bibitem[{Vu et~al.(2020)Vu, Wang, Munkhdalai, Sordoni, Trischler,
  Mattarella-Micke, Maji, and Iyyer}]{vu-etal-2020-exploring}
Tu~Vu, Tong Wang, Tsendsuren Munkhdalai, Alessandro Sordoni, Adam Trischler,
  Andrew Mattarella-Micke, Subhransu Maji, and Mohit Iyyer. 2020.
\newblock \href {https://doi.org/10.18653/v1/2020.emnlp-main.635} {Exploring
  and predicting transferability across {NLP} tasks}.
\newblock In \emph{Proceedings of the 2020 Conference on Empirical Methods in
  Natural Language Processing (EMNLP)}, pages 7882--7926, Online. Association
  for Computational Linguistics.

\bibitem[{Wang et~al.(2019)Wang, Shi, Kim, Oh, Yang, Zhang, and
  Yu}]{wang-etal-2019-persuasion}
Xuewei Wang, Weiyan Shi, Richard Kim, Yoojung Oh, Sijia Yang, Jingwen Zhang,
  and Zhou Yu. 2019.
\newblock \href {https://doi.org/10.18653/v1/P19-1566} {Persuasion for good:
  Towards a personalized persuasive dialogue system for social good}.
\newblock In \emph{Proceedings of the 57th Annual Meeting of the Association
  for Computational Linguistics}, pages 5635--5649, Florence, Italy.
  Association for Computational Linguistics.

\bibitem[{Wang et~al.(2022{\natexlab{a}})Wang, Mishra, Alipoormolabashi, Kordi,
  Mirzaei, Arunkumar, Ashok, Dhanasekaran, Naik, Stap
  et~al.}]{wang2022benchmarking}
Yizhong Wang, Swaroop Mishra, Pegah Alipoormolabashi, Yeganeh Kordi, Amirreza
  Mirzaei, Anjana Arunkumar, Arjun Ashok, Arut~Selvan Dhanasekaran, Atharva
  Naik, David Stap, et~al. 2022{\natexlab{a}}.
\newblock Benchmarking generalization via in-context instructions on 1,600+
  language tasks.
\newblock \emph{arXiv preprint arXiv:2204.07705}.

\bibitem[{Wang et~al.(2022{\natexlab{b}})Wang, Mishra, Alipoormolabashi, Kordi
  et~al.}]{wangmishra2022benchmarkinggeneralization}
Yizhong Wang, Swaroop Mishra, Pegah Alipoormolabashi, Yeganeh Kordi, et~al.
  2022{\natexlab{b}}.
\newblock Benchmarking generalization via in-context instructions on 1,600+
  language tasks.
\newblock \emph{arXiv}.

\bibitem[{Wang et~al.(2020{\natexlab{a}})Wang, Lipton, and
  Tsvetkov}]{wang-etal-2020-negative}
Zirui Wang, Zachary~C. Lipton, and Yulia Tsvetkov. 2020{\natexlab{a}}.
\newblock \href {https://doi.org/10.18653/v1/2020.emnlp-main.359} {On negative
  interference in multilingual models: Findings and a meta-learning treatment}.
\newblock In \emph{Proceedings of the 2020 Conference on Empirical Methods in
  Natural Language Processing (EMNLP)}, pages 4438--4450, Online. Association
  for Computational Linguistics.

\bibitem[{Wang et~al.(2020{\natexlab{b}})Wang, Mehta, Poczos, and
  Carbonell}]{wang-etal-2020-efficient}
Zirui Wang, Sanket~Vaibhav Mehta, Barnabas Poczos, and Jaime Carbonell.
  2020{\natexlab{b}}.
\newblock \href {https://doi.org/10.18653/v1/2020.emnlp-main.39} {Efficient
  meta lifelong-learning with limited memory}.
\newblock In \emph{Proceedings of the 2020 Conference on Empirical Methods in
  Natural Language Processing (EMNLP)}, pages 535--548, Online. Association for
  Computational Linguistics.

\bibitem[{Webson and Pavlick(2021)}]{webson-pavlick-2021}
Albert Webson and Ellie Pavlick. 2021.
\newblock \href {http://arxiv.org/abs/2109.01247} {Do prompt-based models
  really understand the meaning of their prompts?}

\bibitem[{Wei et~al.(2022)Wei, Bosma, Zhao, Guu, Yu, Lester, Du, Dai, and
  Le}]{flan}
Jason Wei, Maarten~Paul Bosma, Vincent Zhao, Kelvin Guu, Adams~Wei Yu, Brian
  Lester, Nan Du, Andrew~Mingbo Dai, and Quoc~V. Le. 2022.
\newblock \href {https://openreview.net/forum?id=gEZrGCozdqR} {Finetuned
  language models are zero-shot learners}.

\bibitem[{Welleck et~al.(2019)Welleck, Weston, Szlam, and
  Cho}]{welleck-etal-2019-dialogue}
Sean Welleck, Jason Weston, Arthur Szlam, and Kyunghyun Cho. 2019.
\newblock \href {https://doi.org/10.18653/v1/P19-1363} {Dialogue natural
  language inference}.
\newblock In \emph{Proceedings of the 57th Annual Meeting of the Association
  for Computational Linguistics}, pages 3731--3741, Florence, Italy.
  Association for Computational Linguistics.

\bibitem[{Weller et~al.(2020)Weller, Lourie, Gardner, and
  Peters}]{weller-etal-2020-learning}
Orion Weller, Nicholas Lourie, Matt Gardner, and Matthew~E. Peters. 2020.
\newblock \href {https://doi.org/10.18653/v1/2020.emnlp-main.105} {Learning
  from task descriptions}.
\newblock In \emph{Proceedings of the 2020 Conference on Empirical Methods in
  Natural Language Processing (EMNLP)}, pages 1361--1375, Online. Association
  for Computational Linguistics.

\bibitem[{Wen et~al.(2017)Wen, Vandyke, Mrk{\v{s}}i{\'c}, Ga{\v{s}}i{\'c},
  Rojas-Barahona, Su, Ultes, and Young}]{wen-etal-2017-network}
Tsung-Hsien Wen, David Vandyke, Nikola Mrk{\v{s}}i{\'c}, Milica
  Ga{\v{s}}i{\'c}, Lina~M. Rojas-Barahona, Pei-Hao Su, Stefan Ultes, and Steve
  Young. 2017.
\newblock \href {https://aclanthology.org/E17-1042} {A network-based end-to-end
  trainable task-oriented dialogue system}.
\newblock In \emph{Proceedings of the 15th Conference of the {E}uropean Chapter
  of the Association for Computational Linguistics: Volume 1, Long Papers},
  pages 438--449, Valencia, Spain. Association for Computational Linguistics.

\bibitem[{Whang et~al.(2021)Whang, Lee, Oh, Lee, Han, Lee, and
  Lee}]{Whang_Lee_Oh_Lee_Han_Lee_Lee_2021}
Taesun Whang, Dongyub Lee, Dongsuk Oh, Chanhee Lee, Kijong Han, Dong-hun Lee,
  and Saebyeok Lee. 2021.
\newblock \href {https://ojs.aaai.org/index.php/AAAI/article/view/17653} {Do
  response selection models really know what’s next? utterance manipulation
  strategies for multi-turn response selection}.
\newblock \emph{Proceedings of the AAAI Conference on Artificial Intelligence},
  35(16):14041--14049.

\bibitem[{Wu et~al.(2020{\natexlab{a}})Wu, Hoi, Socher, and
  Xiong}]{wu-etal-2020-tod}
Chien-Sheng Wu, Steven~C.H. Hoi, Richard Socher, and Caiming Xiong.
  2020{\natexlab{a}}.
\newblock \href {https://doi.org/10.18653/v1/2020.emnlp-main.66} {{TOD}-{BERT}:
  Pre-trained natural language understanding for task-oriented dialogue}.
\newblock In \emph{Proceedings of the 2020 Conference on Empirical Methods in
  Natural Language Processing (EMNLP)}, pages 917--929, Online. Association for
  Computational Linguistics.

\bibitem[{Wu et~al.(2021)Wu, Madotto, Liu, Fung, and Xiong}]{wu2021qaconv}
Chien-Sheng Wu, Andrea Madotto, Wenhao Liu, Pascale Fung, and Caiming Xiong.
  2021.
\newblock Qaconv: Question answering on informative conversations.
\newblock \emph{arXiv preprint arXiv:2105.06912}.

\bibitem[{Wu et~al.(2020{\natexlab{b}})Wu, Zhang, and
  R{\'e}}]{wu2020understanding}
Sen Wu, Hongyang~R Zhang, and Christopher R{\'e}. 2020{\natexlab{b}}.
\newblock Understanding and improving information transfer in multi-task
  learning.
\newblock \emph{arXiv preprint arXiv:2005.00944}.

\bibitem[{Xing et~al.(2022)Xing, Cai, Barlaug, Liu, and
  Gulla}]{xing2022balancing}
Yujie Xing, Jinglun Cai, Nils Barlaug, Peng Liu, and Jon~Atle Gulla. 2022.
\newblock Balancing multi-domain corpora learning for open-domain response
  generation.
\newblock \emph{arXiv preprint arXiv:2205.02570}.

\bibitem[{Xu et~al.(2022)Xu, Chen, Du, Shao, Wang, Li, and
  Yang}]{xu2022zeroprompt}
Hanwei Xu, Yujun Chen, Yulun Du, Nan Shao, Yanggang Wang, Haiyu Li, and Zhilin
  Yang. 2022.
\newblock Zeroprompt: Scaling prompt-based pretraining to 1,000 tasks improves
  zero-shot generalization.
\newblock \emph{arXiv preprint arXiv:2201.06910}.

\bibitem[{Xu et~al.(2021{\natexlab{a}})Xu, Ju, Li, Boureau, Weston, and
  Dinan}]{xu-etal-2021-bot}
Jing Xu, Da~Ju, Margaret Li, Y-Lan Boureau, Jason Weston, and Emily Dinan.
  2021{\natexlab{a}}.
\newblock \href {https://doi.org/10.18653/v1/2021.naacl-main.235}
  {Bot-adversarial dialogue for safe conversational agents}.
\newblock In \emph{Proceedings of the 2021 Conference of the North American
  Chapter of the Association for Computational Linguistics: Human Language
  Technologies}, pages 2950--2968, Online. Association for Computational
  Linguistics.

\bibitem[{Xu et~al.(2021{\natexlab{b}})Xu, Tao, Jiang, Zhao, Zhao, and
  Yan}]{Xu_Tao_Jiang_Zhao_Zhao_Yan_2021}
Ruijian Xu, Chongyang Tao, Daxin Jiang, Xueliang Zhao, Dongyan Zhao, and Rui
  Yan. 2021{\natexlab{b}}.
\newblock \href {https://ojs.aaai.org/index.php/AAAI/article/view/17666}
  {Learning an effective context-response matching model with self-supervised
  tasks for retrieval-based dialogues}.
\newblock \emph{Proceedings of the AAAI Conference on Artificial Intelligence},
  35(16):14158--14166.

\bibitem[{Yang et~al.(2021)Yang, Li, and Quan}]{Yang_Li_Quan_2021}
Yunyi Yang, Yunhao Li, and Xiaojun Quan. 2021.
\newblock \href {https://ojs.aaai.org/index.php/AAAI/article/view/17674} {Ubar:
  Towards fully end-to-end task-oriented dialog system with gpt-2}.
\newblock \emph{Proceedings of the AAAI Conference on Artificial Intelligence},
  35(16):14230--14238.

\bibitem[{Yeh et~al.(2021)Yeh, Eskenazi, and
  Mehri}]{yeh-etal-2021-comprehensive}
Yi-Ting Yeh, Maxine Eskenazi, and Shikib Mehri. 2021.
\newblock \href {https://doi.org/10.18653/v1/2021.eancs-1.3} {A comprehensive
  assessment of dialog evaluation metrics}.
\newblock In \emph{The First Workshop on Evaluations and Assessments of Neural
  Conversation Systems}, pages 15--33, Online. Association for Computational
  Linguistics.

\bibitem[{Yu et~al.(2020)Yu, Sun, Cardie, and Yu}]{yu-etal-2020-dialogue}
Dian Yu, Kai Sun, Claire Cardie, and Dong Yu. 2020.
\newblock \href {https://doi.org/10.18653/v1/2020.acl-main.444} {Dialogue-based
  relation extraction}.
\newblock In \emph{Proceedings of the 58th Annual Meeting of the Association
  for Computational Linguistics}, pages 4927--4940, Online. Association for
  Computational Linguistics.

\bibitem[{Zellers et~al.(2021)Zellers, Holtzman, Clark, Qin, Farhadi, and
  Choi}]{zellers-etal-2021-turingadvice}
Rowan Zellers, Ari Holtzman, Elizabeth Clark, Lianhui Qin, Ali Farhadi, and
  Yejin Choi. 2021.
\newblock \href {https://doi.org/10.18653/v1/2021.naacl-main.386}
  {{T}uring{A}dvice: A generative and dynamic evaluation of language use}.
\newblock In \emph{Proceedings of the 2021 Conference of the North American
  Chapter of the Association for Computational Linguistics: Human Language
  Technologies}, pages 4856--4880, Online. Association for Computational
  Linguistics.

\bibitem[{Zhang et~al.(2021)Zhang, Chen, D{'}Haro, Zhang, Friedrichs, Lee, and
  Li}]{zhang-etal-2021-DynaEval}
Chen Zhang, Yiming Chen, Luis~Fernando D{'}Haro, Yan Zhang, Thomas Friedrichs,
  Grandee Lee, and Haizhou Li. 2021.
\newblock \href {https://doi.org/10.18653/v1/2021.acl-long.441} {{D}yna{E}val:
  Unifying turn and dialogue level evaluation}.
\newblock In \emph{Proceedings of the 59th Annual Meeting of the Association
  for Computational Linguistics and the 11th International Joint Conference on
  Natural Language Processing (Volume 1: Long Papers)}, pages 5676--5689,
  Online. Association for Computational Linguistics.

\bibitem[{Zhang et~al.(2018)Zhang, Dinan, Urbanek, Szlam, Kiela, and
  Weston}]{zhang-etal-2018-personalizing}
Saizheng Zhang, Emily Dinan, Jack Urbanek, Arthur Szlam, Douwe Kiela, and Jason
  Weston. 2018.
\newblock \href {https://doi.org/10.18653/v1/P18-1205} {Personalizing dialogue
  agents: {I} have a dog, do you have pets too?}
\newblock In \emph{Proceedings of the 56th Annual Meeting of the Association
  for Computational Linguistics (Volume 1: Long Papers)}, pages 2204--2213,
  Melbourne, Australia. Association for Computational Linguistics.

\bibitem[{Zhang et~al.(2020)Zhang, Sun, Galley, Chen, Brockett, Gao, Gao, Liu,
  and Dolan}]{zhang-etal-2020-dialogpt}
Yizhe Zhang, Siqi Sun, Michel Galley, Yen-Chun Chen, Chris Brockett, Xiang Gao,
  Jianfeng Gao, Jingjing Liu, and Bill Dolan. 2020.
\newblock \href {https://doi.org/10.18653/v1/2020.acl-demos.30} {{DIALOGPT} :
  Large-scale generative pre-training for conversational response generation}.
\newblock In \emph{Proceedings of the 58th Annual Meeting of the Association
  for Computational Linguistics: System Demonstrations}, pages 270--278,
  Online. Association for Computational Linguistics.

\bibitem[{Zhao et~al.(2020{\natexlab{a}})Zhao, Lala, and
  Kawahara}]{zhao-etal-2020-designing}
Tianyu Zhao, Divesh Lala, and Tatsuya Kawahara. 2020{\natexlab{a}}.
\newblock \href {https://doi.org/10.18653/v1/2020.acl-main.4} {Designing
  precise and robust dialogue response evaluators}.
\newblock In \emph{Proceedings of the 58th Annual Meeting of the Association
  for Computational Linguistics}, pages 26--33, Online. Association for
  Computational Linguistics.

\bibitem[{Zhao et~al.(2020{\natexlab{b}})Zhao, Xu, and
  Wu}]{zhao-etal-2020-learning-simple}
Yufan Zhao, Can Xu, and Wei Wu. 2020{\natexlab{b}}.
\newblock \href {https://doi.org/10.18653/v1/2020.emnlp-main.279} {Learning a
  simple and effective model for multi-turn response generation with auxiliary
  tasks}.
\newblock In \emph{Proceedings of the 2020 Conference on Empirical Methods in
  Natural Language Processing (EMNLP)}, pages 3472--3483, Online. Association
  for Computational Linguistics.

\bibitem[{Zhong et~al.(2021{\natexlab{a}})Zhong, Yin, Yu, Zaidi, Mutuma, Jha,
  Awadallah, Celikyilmaz, Liu, Qiu, and Radev}]{zhong-etal-2021-qmsum}
Ming Zhong, Da~Yin, Tao Yu, Ahmad Zaidi, Mutethia Mutuma, Rahul Jha,
  Ahmed~Hassan Awadallah, Asli Celikyilmaz, Yang Liu, Xipeng Qiu, and Dragomir
  Radev. 2021{\natexlab{a}}.
\newblock \href {https://doi.org/10.18653/v1/2021.naacl-main.472} {{QMS}um: A
  new benchmark for query-based multi-domain meeting summarization}.
\newblock In \emph{Proceedings of the 2021 Conference of the North American
  Chapter of the Association for Computational Linguistics: Human Language
  Technologies}, pages 5905--5921, Online. Association for Computational
  Linguistics.

\bibitem[{Zhong et~al.(2021{\natexlab{b}})Zhong, Lee, Zhang, and
  Klein}]{zhong-etal-2021-adapting-language}
Ruiqi Zhong, Kristy Lee, Zheng Zhang, and Dan Klein. 2021{\natexlab{b}}.
\newblock \href {https://doi.org/10.18653/v1/2021.findings-emnlp.244} {Adapting
  language models for zero-shot learning by meta-tuning on dataset and prompt
  collections}.
\newblock In \emph{Findings of the Association for Computational Linguistics:
  EMNLP 2021}, pages 2856--2878, Punta Cana, Dominican Republic. Association
  for Computational Linguistics.

\end{thebibliography}
\bibliographystyle{acl_natbib}

\section*{Appendix}
\label{sec:appendix}

\renewcommand{\thesubsection}{\Alph{subsection}}

\subsection{Additional implementation details}
\label{sec:extraimplementation}
\noindent
\textbf{Data Sampling} 
For training data creation, we first generate instances from all datasets belonging to each task. Since the number of instances per task can be highly imbalanced, we sample a fixed maximum of $N$ number of instances per task. In our main models and experiments, we set $N=5000$. Each instance in a task is assigned a random task definition and prompt. We truncate the input sequences to 1024 tokens and target output sequences to 256 tokens.

\noindent
\textbf{Implementation Details}
Our models are trained for 3 epochs with a learning rate of 5e-5 with an Adam optimizer~\cite{kingma:adam} with linear learning rate decay. For our main experiments in Table~\ref{tab:mainresults}, we perform checkpoint selection using a validation set created from the train tasks. For rest of the experiments we do model selection using the validation sets.
We use the HuggingFace Transformers library\footnote{\url{https://github.com/huggingface/transformers}} for training and inference implementation and use Deepspeed library\footnote{\url{https://github.com/microsoft/DeepSpeed}} for improving training efficiency. We train \modelnamebartzero on 2 Nvidia 2080Ti GPUs using a batch size of 2 per GPU and an effective batch size of 72 with gradient checkpointing. We train \modelnametzero on 2 Nvidia A6000 GPUs using a batch size of 1 per GPU and an effective batch size of 72 with gradient checkpointing. For all classification tasks, we perform greedy decoding, and for all generation tasks, we perform top-p sampling with $p=0.7$ and temperature set to 0.7. The repetition penalty is set to 1.2. In Table~\ref{tab:mainresults}, for \modelnamebartzero and \modelnametzero, we report the results over three different training runs, where each run is based on a new sample of training data. 

\noindent
\textbf{Zero-shot Automatic Evaluation Implementation Details}
For zero shot automatic evaluation, we calculate the Spearman correlation of the model's prediction with human ratings for relevance provided in the DSTC-10 test sets. There is no consistent ``relevance'' or ``coherence'' rating field present across the evaluation datasets. We therefore calculate the correlation with the ratings if a rating exists in any of the following fields ``overall'', ``turing'', ``relevance'' and ``appropriateness''.

\begin{table*}[]
\centering
\small
\begin{tabular}{p{0.18\linewidth}p{0.5\linewidth}p{0.26\linewidth}}
\toprule
\multicolumn{3}{p{0.9\linewidth}}{Conversation:}                    \\
\multicolumn{3}{p{0.96\linewidth}}{{[}CONTEXT{]} How may I help you? {[}ENDOFTURN{]} }                    \\
\multicolumn{3}{p{0.96\linewidth}}{ I left a suitcase on the train to London the other day. {[}ENDOFDIALOGUE{]}}                    \\
\midrule
\textbf{Task}     & \textbf{Instruction} & \textbf{Output}                                                             \\
\midrule
Response editing                                                                       & Modify the provided response into a response that is fluent and coherent to the dialogue context: {[}RESPONSE{]} Can describe it it , sir ? It will help us find              & Can you describe it, sir? It will help us find it.                 \\
\midrule
Begins with                                                                            & Generate a response that starts with the provided initial phrase.  {[}INITIAL PHRASE{]} Please describe                                                                       & Please describe the suitcase.                                      \\
\midrule
Begins with + Keyword controlled generation & Generate a response that starts with the provided initial phrase and contains the provided keywords. {[}INITIAL PHRASE{]} Please describe {[}KEYWORDS{]} color, any documents & Please describe the color of the suitcase and any documents in it. \\
\midrule
Intent detection                                                                       & What is the intent of the response {[}OPTIONS{]} booking, reservation change, checkout, lost\&found,..., time information, security, schedules                                & lost\&found                                                        \\
\midrule
Summarization                                                                          & Return a summary of the provided conversation.                                                                                                                          & Person2 left a suitcase on the train to London the other day.      \\
\midrule
Answer generation                                                                      & {[}QUESTION{]} What is the response of following question:  Where was the person going to?                                                                                     & London                                                             \\
\midrule
Knowledge grounded generation                                                          & Generate a response using the provided background knowledge. {[}KNOWLEDGE{]} Emailid for cases related to lost\&found is x@gmail.com                                          & You can contact us at x@gmail.com \\
\bottomrule
\end{tabular}
\caption{A sample conversation followed by instructions for multiple tasks for that conversation, and the outputs generated based on the specified instructions. Instruction tuning allows performing multiple tasks on an input by specifying task-specific instructions and prompts.}
\label{tab:exampleoutputs}
\end{table*}

\subsection{Sample conversation and Instructions}
\label{sec:sampleconversation}
In Table~\ref{tab:exampleoutputs}  we provide a sample conversation followed by instructions for multiple tasks for that conversation, and the outputs generated by \modelnamebartzero based on the specified instructions. Through this example we illustrate that instruction tuning allows performing multiple tasks on an input by specifying task-specific instructions.

\subsection{Datasets used in tasks}
In Table \ref{tab:tasksdatasets} we present the list of tasks with datasets used in each task. 

\subsection{Configuration of experiments}
In Table \ref{tab:datasets} we provide the configurations of experiments, that is, the tasks used for training for each experiment.

\begin{table*}[h]
\tiny
    \centering
    \begin{tabular}{l|l|p{0.64\textwidth}}
    \hline
   \textbf{Task Type} & \textbf{Task Name} & \textbf{Example} \\
    \hline
         \multirow{2}{*}{Intent} & Intent Classification & [RES] Make a reservation for 4 [EOD]. The possible intents are: [OPT] BookRestaurant | ShareETA  [Q] The best option is\\
         \cline{2-3}
         & Intent Present &  [RES] list the three earliest flights [EOD]. The possible options are: [OPT] yes | no [Q]. Is the intent flight correct?. " \\
    \hline
         \multirow{3}{*}{NLI} & DialFact &  [CTX] What year did they start making pasta? [RES] I think pasta was first made somewhere in Europe many centuries ago. [EOD] The possible classes are: [OPT] refutes | supports | not enough info [Q]. Choose the most possible class \\
             \cline{2-3}
             & DNLI & [CTX] i am named after a cartoon fox . [RES] i have a dog . [EOD]. The possible classes are: positive | negative | neutral  [Q]. The predicted class is \\
             \cline{2-3}
             & Contradiction & [CTX] lol are they fast drying ? [EOT] Kind of slow lol. [RES] I know they dry fast. [EOD]. The possible classes are: uncontradicted | contradicted [Q]. What is the class given the context and the response \\
    \hline
         Safety Classification & Toxicity Classification & [CTX] Hello [EOT] hello ... [EOD] [REP] not interesting [Q] Is the response toxic? Answer choices [OPT] yes | no \\
    \hline    
        \multirow{2}{*}{Relation Extraction} & Relation Classification & [CTX]It's like this, me, no jokes. [EOT] All right ... [EOD]. The possible relations are: [OPT] per:siblings ... [Q]. The relation between A and B is \\
          \cline{2-3}
            & Relation Present & [CTX] Hello, Mark? ... [EOT]  That is so made up!  [EOD] . Does the relation per:alternate exist between A and B? Answer [OPT] yes | no.\\
    \hline
        \multirow{3}{*}{Evaluation} & Relevance &   [CTX] to holden my dad ... [EOD] [REP] you can send us your email address. [SEP] Is the response contextual? Answer [OPT] yes | no.\\
          \cline{2-3}
            & Selection & [CTX] this is sprint great service URL [EOD] The best response is [OPT] you can send us | please ... \\
          \cline{2-3}
            & Rating & [CTX] this is sprint great service URL [EOD] Please give a rating ranging from 1 to 5 to the following response: please dm us your account \\
    \hline 
        \multirow{3}{*}{Slot} & Slot Classifcation & [RES] what do you have tomorrow after 5 o'clock from atlanta to san francisco [EOD] [Q] What is the value of slot: city\_name in the response  \\
          \cline{2-3}
            & Slot Present & [RES] Yes. That sounds great. Can I scheduled ... [EOD]. The possible options are: [OPT] yes | no [Q]. The slot visit date is present in the response? \\
            \cline{2-3}
            & Slot Value Generation & [CTX] I need tickets to [EOT] Great! [RES] You've got 2 tickets [EOD] [Q]. What is the value of slot: starttime in the response\\
    \hline
        \multirow{2}{*}{Safety Generation} & Non-Toxic Feedback & [CTX] I have never met [EOT] another group is ... [EOD] [Q] Given the conversation, a non toxic response is \\
        \cline{2-3}
            & Recovery Resp. Generation & [CTX] I have never met [EOT] another group is ... [EOD] [Q] Given the conversation, a non toxic recovery response is\\
    \hline 
        \multirow{7}{*}{Grounded Generation} & Emotion & [EMO] anger [CTX] I won! [EOD] [Q] Given the context and emotion, the response is\\
        \cline{2-3}
            & DB based & [STATE] hotelparking: yes [DB] Type: guest house [CTX] there are ... [EOD] [Q] Given the context, db, and state, the response is \\
              \cline{2-3}
            & Document-grounded & [WIKI] you must report .. [CLASS] That is the case ... [CTX] Hello ... [EOD] [Q] Given the context and doc, the response is  \\
              \cline{2-3}
            & Graph Based & [GRAPH] the subject is, relation: [CTX] do you like iron man [EOD] [Q] Given the context and triplets, the response is\\
              \cline{2-3}
            & Persona & [P] i'm 60 years old ... [CTX] Hello! How is your ... [EOD] Given the context and persona, the response is\\
              \cline{2-3}
            & Schema Based & [SCHEMA] terminal: false, label: open circuit [CTX] My car is ... [EOD] [Q] Given this context and schema, the response is \\
              \cline{2-3}
            & Knowledge-Grounded & [DOC] demetri martin was accepted into harvard law , but left out of boredom to pursue a career in comedy [CTX] do you know who demetri martin is ? [EOD] Given this context and knowledge, the response is  \\
    \hline
        \multirow{4}{*}{QA and Commensense} & Answer Generation &[DOC] Jessica went to sit in her rocking chair ... [Q] Who had a Birthday? Jessica. How old would she be? \\
         \cline{2-3}
         & Answer Selection &  [DOC] Jessica went to sit in her r ... [OPT] 80 | park ... [Q] Who had a Birthday? Jessica. How old would she be? \\
          \cline{2-3}
            & Question Generation &  [DOC] Jessica went to sit in her rocking chair. Today was her birthday and she was turning 80 [Q] what should we ask about this conversation \\
             \cline{2-3}
            & Target Guided & [Target] i love chocolate. [CTX] i love walking in the park. [Q] Generate a text which connects the context with the target sentence."\\
    \hline 
        \multirow{4}{*}{Controlled Generation} & Begins With & [INIT] I tell ya [CTX] can I ask you something? ... [EOD] [Q] Given this context generate a response which starts with the given initial sentence: \\
         \cline{2-3}
            & Ends with & [FINAL] checks ? [CTX] Are you through with your meal ... [EOD] [Q] Given this context and final phrase, the response is \\
             \cline{2-3}
            & Keyword Based & [KEY] lot of memory, desktop computer and memory [CTX] Can I help you ... [EOD] [Q] Here is a response which contains the given keywords \\
             \cline{2-3}
            & N Words & [CTX] Do you know Manchester United F.C ... [EOD] [Q] Given this context, the response with 3 number of words is\\
    \hline 
        Dialog State Generation & Dialog State Generation & [CTX] I need help finding an apartment [EOT] what area are you hoping ... [EOD] [Q] What is the belief state? \\
    \hline 
        \multirow{3}{*}{Edit Generation} & Shuffling &  \multirow{3}{*}{[RES] hi, report  [CTX] Many DMV  ... [EOD] [Q] Given this context and response provided, the edited response is}\\
         \cline{2-2}
            & Adding &  \\
             \cline{2-2}
            & Removing &\\
    \hline 
        \multirow{4}{*}{Pretrain Tasks} & Fill Missing Utterance & [CTX] Do you know Manchester United F.C.? ... [EOD] [Q] Given this context generate a response coherent to the context\\
         \cline{2-3}
            & Find Incoherence Utterance & [CTX] Do you know Manchester United F.C.? ... [EOD] [Q] Given this context generate a response coherent to the context \\
             \cline{2-2}
            & Find Missing Utterance & [CTX] Do you know Manchester United F.C.? [EOT] [MASK] ... [EOD] [Q] Here is the missing utterance that can take place of [MASK]\\
             \cline{2-2}
            & Find Swapped Utterance & [CTX] Do you know Manchester United F.C.?  [EOD] [Q] Given this context the swapped indices of responses are\\
            
    \hline 
        \multirow{2}{*}{Response Generation} & Open Domain &  \multirow{2}{*}{[CTX] Do you know Manchester United F.C.? ... [EOD] [Q] Given this context generate a response coherent to the context"} \\
            & Task-oriented &  \\
    \hline
        Summarization & Summary Generation  & [CTX] Person2 OK. [EOT] Person1: Well, how old are you? ... [EOD] [Q] Given this dialog context, its summary is the following: \\
    \hline 
        \multirow{9}{*}{Misc} & Act Classification & [CTX] Hi, I am looking for a nice German restaurant [EOD] The possible acts are: [OPT] request | inform [Q] The dialog act is\\
         \cline{2-3}
            & Advice Present & [CTX] Anyone take mental ... [EOD] [RES] Back at my old job ... [Q] Does the response provide advice for the issue? Choices [OPT] yes | no \\
             \cline{2-3}
              & Advice Generation & [CTX] Anyone take mental health days from work? ... [EOD] [Q] The response is\\
               \cline{2-3}
            & Deal Present & [CTX] I like the basketball and the hat ... [EOT] deal [EOD] [Q] Was an agreement reached? Choices [OPT] yes | no \\
             \cline{2-3}
            & Emotion Tagging & [CTX] Hey, so did you have fun with Joey ... [EOD] The possible emotions are [OPT] disgust ... [Q] The emotions in the dialog are \\
             \cline{2-3}
            & Persuasion Present & [CTX] Hello How are you ...[EOD] [RES] Are you involved with charities [Q] Is task-related-inquiry used in the response? Choices [OPT] yes | no \\
             \cline{2-3}
            & Persuasion Strategy & [CTX] how can i help [EOD] The possible strategies are: [OPT] request | inform [Q] The strategy is \\
             \cline{2-3}
            & Persuasion Generation & [STRATEGY] proposition-of-donation [CTX] how can i help? [EOD] [Q] The response is  \\
             \cline{2-3}
            & Count Response Words & [CTX] Do you know Manchester United F.C.? ... [EOD] [Q] Given this context Here is length of the response in the context"
            \\
    \hline    
    \end{tabular}
    \caption{List of tasks with sample inputs for each task. The left column describes the general task type. The middle column lists the specific task. The right column displays an example formatted using a randomly selected task definition and prompt for the task. [CTX] is short for [CONTEXT], [Q] is short for [QUESTION], [RES] is short for response, [EOT] is short for [ENDOFTURN] and [EOD] is short for [ENDOFDIALOGUE]}
    \label{tab:prompttasks}
    
\end{table*}

\begin{table*}[]
\tiny
    \centering
    \begin{tabular}{l|l|p{0.64\textwidth}}
    \hline
   \textbf{Task Type} & \textbf{Task Name} & \textbf{Datasets} \\
    \hline
         \multirow{2}{*}{Intent} & Intent Classification & ATIS~\cite{hemphill-etal-1990-atis} SNIPS~\cite{coucke2018snips} CLINIC150~\cite{larson-etal-2019-evaluation}\\
         \cline{2-2}
         & Intent Present &    HWU64~\cite{liu2021benchmarking} Banking77~\cite{casanueva-etal-2020-efficient}  \\
    \hline
         \multirow{3}{*}{NLI} & DialFact & DialFact~\cite{gupta2021dialfact} \\
             \cline{2-3}
             & DNLI & \multirow{2}{*}{Decode~\cite{nie-etal-2021-like} Dialogue NLI~\cite{welleck-etal-2019-dialogue}}\\
             \cline{2-2}
             & Contradiction \\
    \hline
         Safety Classification & Toxicity Classification & ToxiChat~\cite{baheti-etal-2021-just} BAD~\cite{xu-etal-2021-bot} Build it Break it Fix it~\cite{dinan-etal-2019-build} \\
    \hline    
        \multirow{2}{*}{Relation Extraction} & Relation Classification & \multirow{2}{*}{DialogRE~\cite{yu-etal-2020-dialogue}} \\
          \cline{2-2}
            & Relation Present \\
    \hline
        \multirow{3}{*}{Evaluation} & Relevance &   DSTC6 \cite{DSTC6_End-to-End_Conversation_Modeling} DSTC7~\cite{Galley2019GroundedRG} Persona-Chatlog~\cite{see-etal-2019-makes}\\
          \cline{2-2}
            & Selection & USR~\cite{mehri-eskenazi-2020-usr} FED~\cite{mehri-eskenazi-2020-unsupervised} DailyDialog~\cite{gupta-etal-2019-[investigating, zhao-etal-2020-designing} \\
          \cline{2-2}
            & Rating & PersonaChat~\cite{zhao-etal-2020-designing} GRADE~\cite{huang-etal-2020-grade} HUMOD~\cite{merdivan2020human}\\
    \hline 
        \multirow{3}{*}{Slot} & Slot Classifcation & RESTAURANTS-8K~\cite{coope-etal-2020-span} DSTC8-SGD~\cite{rastogi2020towards}  \\
          \cline{2-2}
            & Slot Present & ATIS~\cite{hemphill-etal-1990-atis} SNIPS~\cite{coucke2018snips} \\
            \cline{2-2}
            & Slot Value Generation & TaskMaster~\cite{byrne-etal-2019-taskmaster} MSRE2E~\cite{li2018microsoft}\\
    \hline
        \multirow{2}{*}{Safety Generation} & Non-Toxic Feedback & \multirow{2}{*}{SaFeRDialogues~\cite{ung2021saferdialogues}}\\
        \cline{2-2}
            & Recovery Response Generation \\
    \hline 
        \multirow{7}{*}{Grounded Generation} & Emotion & EmpatheticDialogues~\cite{rashkin-etal-2019-towards} GoEmotions~\cite{demszky-etal-2020-goemotions} EmotionLines~\cite{hsu-etal-2018-emotionlines}\\
        \cline{2-3}
            & DB based & MultiWOZ~\cite{budzianowski-etal-2018-multiwoz} \\
              \cline{2-3}
            & Document-grounded & doc2dial~\cite{feng-etal-2020-doc2dial} \\
              \cline{2-3}
            & Graph Based & OpenDialKG~\cite{moon-etal-2019-opendialkg} \\
              \cline{2-3}
            & Persona & ConvAI~\cite{dinan2019second} PersonaChat~\cite{zhang-etal-2018-personalizing}\\
              \cline{2-3}
            & Schema Based & FloDial~\cite{raghu-etal-2021-end} \\
              \cline{2-3}
            & Knowledge-Grounded & TopicalChat~\cite{Gopalakrishnan2019} WoW~\cite{dinan2018wizard} \\
    \hline
        \multirow{4}{*}{QA and Commensense} & Answer Generation & CIDEr~\cite{Vedantam_2015_CVPR} TIMEDIAL~\cite{qin-etal-2021-timedial} MuTual~\cite{cui-etal-2020-mutual}\\
         \cline{2-2}
         & Answer Selection & QAConv~\cite{wu2021qaconv} CoQA~\cite{reddy-etal-2019-coqa} QuAC~\cite{choi-etal-2018-quac} \\
          \cline{2-3}
            & Question Generation & QAConv~\cite{wu2021qaconv} \\
             \cline{2-3}
            & Target Guided & OTTers~\cite{sevegnani-etal-2021-otters} \\
    \hline 
    
        \multirow{4}{*}{Controlled Generation} & Begins With & \multirow{2}{*}{EmpatheticDialogues~\cite{rashkin-etal-2019-towards} DailyDialog~\cite{li-etal-2017-dailydialog} ConvAI~\cite{dinan2019second} } \\
         \cline{2-2}
            & Ends with & \\
             \cline{2-2}
            & Keyword Based & TuringAdvice~\cite{zellers-etal-2021-turingadvice} EmotionLines~\cite{hsu-etal-2018-emotionlines} WoW~\cite{dinan2018wizard}\\
             \cline{2-3}
            & N Words & DailyDialog~\cite{li-etal-2017-dailydialog} WoW~\cite{dinan2018wizard} EmpatheticDialogues~\cite{rashkin-etal-2019-towards} \\
    \hline 
        \multirow{2}{*}{Dialog State Generation} & \multirow{2}{*}{Dialog State Generation} & MultiWOZ~\cite{budzianowski-etal-2018-multiwoz}  KVRET \cite{eric-etal-2017-key} WOZ \cite{mrksic-etal-2017-neural} CamRest676 \cite{wen-etal-2017-network} \\
        & & MSR-E2E \cite{li2018microsoft} Frames \cite{el-asri-etal-2017-frames} TaskMaster \cite{byrne-etal-2019-taskmaster} Schema-Guided \cite{rastogi2020towards} \\
    \hline 

        \multirow{3}{*}{Edit Generation} & Shuffling & TopicalChat~\cite{Gopalakrishnan2019} EmotionLines~\cite{hsu-etal-2018-emotionlines}
        EmpatheticDialogues~\cite{rashkin-etal-2019-towards} \\
         \cline{2-2}
            & Adding & WoW~\cite{dinan2018wizard} Persuation~\cite{wang-etal-2019-persuasion} CaSiNo~\cite{chawla-etal-2021-casino} DialogSum~\cite{chen-etal-2021-dialogsum} \\
             \cline{2-2}
            & Removing & DailyDialog~\cite{li-etal-2017-dailydialog} ConvAI~\cite{dinan2019second} EmotionLines~\cite{hsu-etal-2018-emotionlines} \\
    \hline 
        \multirow{4}{*}{Pretrain Tasks} & Fill Missing Utterance & \multirow{4}{*}{DailyDialog~\cite{li-etal-2017-dailydialog} WoW~\cite{dinan2018wizard} EmpatheticDialogues OpenDialKG~\cite{moon-etal-2019-opendialkg}  }\\
         \cline{2-2}
            & Find Incoherence Utterance \\
             \cline{2-2}
            & Find Missing Utterance \\
             \cline{2-2}
            & Find Swapped Utterance \\
            
    \hline 
    
        \multirow{3}{*}{Response Generation} & \multirow{2}{*}{Open Domain} & DailyDialog~\cite{li-etal-2017-dailydialog}  ConvAI~\cite{dinan2019second} WoW~\cite{dinan2018wizard} \\
        & & EmpatheticDialogues~\cite{rashkin-etal-2019-towards} OpenDialKG~\cite{moon-etal-2019-opendialkg}   \\
        \cline{2-3}
            & Task-oriented & MultiWOZ~\cite{budzianowski-etal-2018-multiwoz} \\
    \hline
        Summarization & Summary Generation  & DialSum~\cite{goo2018abstractive} QMSum~\cite{zhong-etal-2021-qmsum} SAMSum~\cite{gliwa2019samsum} \\
    \hline 
        \multirow{9}{*}{Misc} & Act Classification & MSRE2E~\cite{li2018microsoft} DailyDialog~\cite{li-etal-2017-dailydialog}  MultiWOZ~\cite{budzianowski-etal-2018-multiwoz} \\
         \cline{2-3}
            & Advice Present & \multirow{2}{*}{TuringAdvice~\cite{zellers-etal-2021-turingadvice}}\\
             \cline{2-2}
              & Advice Generation \\
               \cline{2-3}
            & Deal Present & Deal~\cite{lewis-etal-2017-deal} \\
             \cline{2-3}
            & Emotion Tagging & GoEmotions~\cite{demszky-etal-2020-goemotions} EmotionLines~\cite{hsu-etal-2018-emotionlines} DailyDialog~\cite{li-etal-2017-dailydialog} \\
             \cline{2-3}
            & Persuasion Present & \multirow{3}{*}{Persuation~\cite{wang-etal-2019-persuasion} CaSiNo~\cite{chawla-etal-2021-casino}}\\
             \cline{2-2}
            & Persuasion Strategy \\
             \cline{2-2}
            & Persuasion Generation \\
             \cline{2-3}
            & Count Response Words & DailyDialog~\cite{li-etal-2017-dailydialog} WoW~\cite{dinan2018wizard} EmpatheticDialogues~\cite{rashkin-etal-2019-towards}
            \\
    \hline    
    \end{tabular}
    \caption{List of Tasks with datasets used in each task. The left column describes the general task type. The middle column lists the specific task. The right column shows all datasets used for a specific task type.}
    \label{tab:tasksdatasets}
    
\end{table*}

\begin{table*}
    \centering
    \tiny
    \begin{tabular}{p{2cm}|p{2cm}|p{3cm}p{3cm}p{3cm}}
    \hline
    \textbf{Experiment} & \textbf{Base model(s)} & \textbf{Tasks} & & \\
    \hline
    Main zero-shot tasks & ID-BART0, ID-T0 & act classification
					\newline act generation
					\newline advice generation
					\newline advice present
					\newline answer generation
					\newline count response words
					\newline db based generation
					\newline deal present
					\newline document grounded generation
					\newline edit generation
					\newline emotion generation
					\newline emotion tagging
					\newline endswith controlled generation
                    & 
                    fill missing utterance
					\newline find incoherent utterance
					\newline find missing utterance
					\newline graph based generation
					\newline intent classification
					\newline intent present (no intent banking dataset)
					\newline keyword controlled generation
					\newline nli classification
					\newline nontoxic feedback generation
					\newline persona grounded generation
					\newline persuasion generation
                    & 
                    persuasion present
					\newline persuasion strategy
					\newline question generation
					\newline recovery generation
					\newline response generation
					\newline response generation with n words
					\newline schema based generation
					\newline slot present
					\newline slot value generation
					\newline summarization
					\newline target controlled generation
					\newline toxic classification \\
    \hline
    Evaluation & ID-BART0  & act classification
                    \newline act generation
                    \newline advice present
                    \newline answer generation
                    \newline answer selection
                    \newline beginswith controlled generation
                    \newline belief state generation
                    \newline db based generation
                    \newline deal present
                    \newline document grounded generation
                    \newline emotion generation
                    & 
                    emotion tagging
                    \newline endswith controlled generation
                    \newline graph based generation
                    \newline intent classification
                    \newline intent present 
                    \newline keyword controlled generation
                    \newline knowledge grounded generation
                    \newline nli classification
                    \newline persona grounded generation
                    \newline persuasion generation
                    \newline persuasion present
                    & 
                    question generation
                    \newline relation classification
                    \newline relation present
                    \newline response generation
                    \newline schema based generation
                    \newline slot present
                    \newline slot value generation
                    \newline summarization
                    \newline target controlled generation \\
    \hline
    Dialog State Generation & ID-BART0 & act classification
                    \newline act generation
                    \newline advice generation
                    \newline advice present
                    \newline answer generation
					\newline answer selection
					\newline beginswith controlled generation
					\newline count response words
					\newline db based generation
					\newline deal present
					\newline dialfact classification
					\newline dialog state generation (no multiwoz)
					\newline document grounded generation
					\newline edit generation
					\newline emotion generation
                    & 
                    emotion tagging
					\newline endswith controlled generation
					\newline fill missing utterance
					\newline find incoherent utterance
					\newline find missing utterance
					\newline find swapped utterance
					\newline gensf slot tagging
				    \newline graph based generation
					\newline intent classification
					\newline intent present
					\newline keyword controlled generation
					\newline knowledge grounded generation
					\newline nli classification
					\newline nontoxic feedback generation
					\newline persona grounded generation
                    & 
                    persuasion generation
					\newline persuasion present
					\newline persuasion strategy
					\newline question generation
					\newline recovery generation
					\newline relation classification
					\newline relation present
					\newline response generation
					\newline response generation with n words
					\newline schema based generation
					\newline slot present
					\newline slot value generation
					\newline summarization
					\newline target controlled generation
					\newline toxic classification \\
    \hline
    Slot Filling & ID-BART0  & act classification
					\newline act generation
					\newline answer generation
					\newline answer selection
					\newline beginswith controlled generation
					\newline belief state generation
					\newline count response words
					\newline db based generation
					\newline deal present
					\newline dialfact classification
					\newline document grounded generation
					\newline edit generation
					\newline emotion generation
					\newline emotion tagging
					\newline endswith controlled generation
                    & 
                    eval binary
					\newline eval ranking
					\newline eval rating
					\newline fill missing utterance
					\newline find incoherent utterance
					\newline find missing utterance
					\newline find swapped utterance
					\newline intent classification
					\newline intent present
					\newline keyword controlled generation
					\newline knowledge grounded generation
					\newline nli classification
					\newline nontoxic feedback generation
					\newline persona grounded generation
					\newline persuasion generation
                    & 
                    persuasion present
					\newline persuasion strategy
					\newline question generation
					\newline recovery generation
					\newline relation classification
					\newline relation present
					\newline response generation
					\newline response generation with n words
					\newline schema based generation
					\newline slot present
					\newline slot value generation
					\newline summarization
					\newline target controlled generation
					\newline toxic classification \\
	\hline	
    \end{tabular}
    \caption{List of experiments and their base models. The tasks listed in the right column are all the tasks a base model was trained with for their corresponding experiment.}
    \label{tab:datasets}
\end{table*}

\end{document}